\documentclass[10pt, final,journal,twocolumn,letterpaper,oneside]{IEEEtran}
\usepackage{subfigure}
\usepackage{times}
\usepackage{url}
\usepackage{bm}
\usepackage[colorlinks,linkcolor=blue]{hyperref}
\usepackage{amsmath,amssymb,amsfonts,mathrsfs,bm}
\newtheorem{theorem}{Theorem}

\usepackage{caption}
\usepackage{graphicx}
\usepackage{graphicx,epsfig,algorithmic,algorithm}
\usepackage{pifont}
\usepackage{array}
\usepackage{multirow}
\usepackage{lscape} 
\usepackage{float}
\usepackage{booktabs,diagbox}
\usepackage{enumerate}
\usepackage{setspace}
\usepackage{epstopdf}
\usepackage{threeparttable}
\usepackage[T1]{fontenc}
\usepackage[table]{xcolor}
\usepackage{collcell}
\usepackage{hhline}
\usepackage{pgf}

\def\colorModel{hsb} 

\newcommand\ColCell[1]{
	\pgfmathparse{#1<50?0:1}  
	\ifnum\pgfmathresult=0\relax\color{white}\fi
	\pgfmathsetmacro\compA{0}      
	\pgfmathsetmacro\compB{0} 
	\pgfmathsetmacro\compC{#1/100}      
	\edef\x{\noexpand\centering\noexpand\cellcolor[\colorModel]{\compA,\compB,\compC}}\x #1
} 
\newcolumntype{E}{>{\collectcell\ColCell}m{0.6cm}<{\endcollectcell}}  

\newenvironment{proof}{{\noindent\it\textbf{Proof }}}{\hfill $\square$\par}

\newcommand{\Rmnum}[1]{\expandafter\@slowromancap\romannumeral #1@}
\makeatother

\begin{document}
	
	\title{A Deeper Look at Facial Expression Dataset Bias}
	\author{Shan~Li, 
		and~Weihong~Deng$^*$,~\IEEEmembership{Member,~IEEE}
		\IEEEcompsocitemizethanks{\IEEEcompsocthanksitem The authors are with the Pattern Recognition and Intelligent System Laboratory, School of Information and Communication Engineering, Beijing University of Posts and Telecommunications, Beijing, 100876, China. \protect\\
			E-mail:\{ls1995, whdeng\}@bupt.edu.cn.}
	}
	\maketitle

	\begin{abstract}
		Datasets play an important role in the progress of facial expression recognition algorithms, but they may suffer from obvious biases caused by different cultures and collection conditions. To look deeper into this bias, we first conduct comprehensive experiments on dataset recognition and cross-dataset generalization tasks, and for the first time explore the intrinsic causes of the dataset discrepancy. The results quantitatively verify that current datasets have a strong build-in bias and corresponding analyses indicate that the conditional probability distributions between source and target datasets are different. However, previous researches are mainly based on shallow features with limited discriminative ability under the assumption that the conditional distribution remains unchanged across domains.
		To address these issues, we further propose a novel deep Emotion-Conditional Adaption Network (ECAN) to learn domain-invariant and discriminative feature representations, which can match both the marginal and the conditional distributions across domains simultaneously. In addition, the largely ignored expression class distribution bias is also addressed by a learnable re-weighting parameter, so that the training and testing domains can share similar class distribution. Extensive cross-database experiments on both lab-controlled datasets (CK+, JAFFE, MMI and Oulu-CASIA) and real-world databases (AffectNet, FER2013, RAF-DB 2.0 and SFEW 2.0) demonstrate that our ECAN can yield competitive performances across various facial expression transfer tasks and outperform the state-of-the-art methods.
		
	\end{abstract}
	
	
	
		\section{Introduction}
	\label{sec:intro}
	\IEEEPARstart{A}{utomatic} facial expression recognition (FER) has been one of the research hotspots in computer vision and machine learning for its huge potential applications in human-computer interface (HCI). Early researches in facial expression recognition mainly evaluate within a single facial expression database and can achieve promising performances in controlled laboratory environment \cite{shan2009facial,littlewort2006dynamics}. However, lab-controlled databases with posed expressions are too uniform to reflect complex scenarios in our real life. In this context, more and more datasets have been collected from real world for conducting facial expression analysis under unconstrained and challenging conditions, which contain more factors unrelated to facial expressions, such as registration errors and variations in pose, occlusion, illumination, background and subject identity \cite{dhall2011static,li2017reliable}.  With this development of diverse facial expression datasets, the dataset bias problem becomes obvious, especially between lab-controlled and real-world data collections. Moreover, in practical applications testing face images could be taken from varying domains. Hence it is hard to satisfy the assumption that the training (source) dataset and test (target) dataset share an identical distribution. Although current research has realized the existence of facial expression dataset bias, it still lacks in-depth study on how  exactly these facial expression datasets differ from each other and in what aspects the bias actually affects the classification performance.
	
	For a deeper look at dataset bias, we conduct a series of comprehensive experiments on popular facial expression datasets  and provide a reliable measure to evaluate the dataset bias quantitatively.
	In the \textit{data recognition experiment},  the best classification performance on seven facial expression datasets reaches 79\% despite the small training sample size (the chance is 1/7=14\%), and intriguingly, there is no evidence of saturation as more training data is added. And the \textit{cross-dataset generalization experiment} on specific expression categories also reveals that facial expression datasets appear to have a strong build-in bias and even the deep models are insufficient to generalize well across various datasets.  
	
		\begin{figure}[t]
		\centering
		\includegraphics[width=8.5cm]{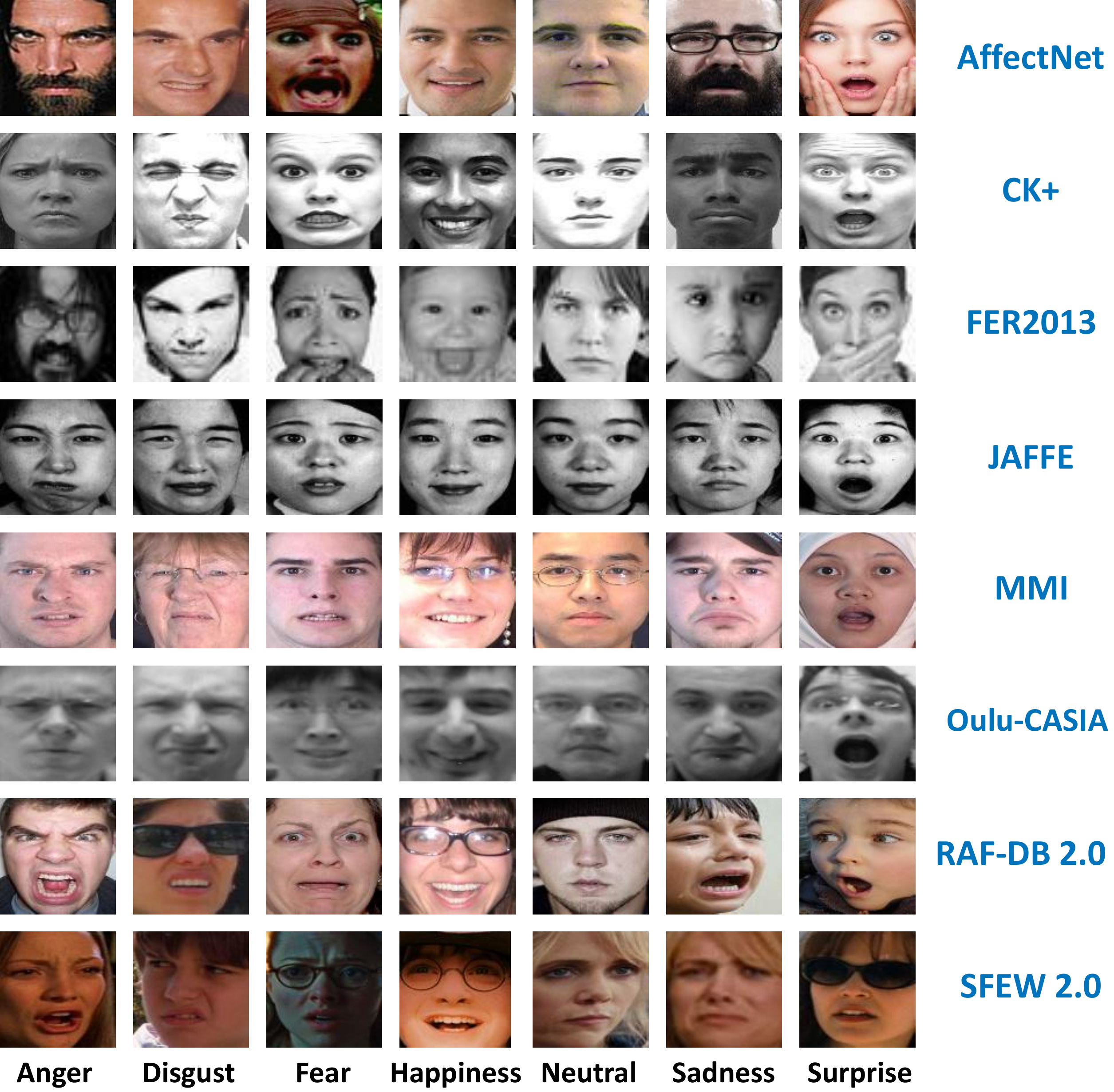}
		\caption{ Already aligned images from different facial expression datasets. Expression label, from left to right, is Anger, Disgust, Fear, Happiness, Neutral, Sadness, Surprise. Even after aligning and eliminating background variations, domain discrepancy still lingers  among these facial expression datasets due to different settings on people's age range, gender, culture, pose and the level of expressiveness.}
		\label{fig:bias}
	\end{figure}

	
	Focusing on the above challenges, we propose a novel deep Emotion-Conditional Adaption Network (ECAN) for \textit{unsupervised cross-dataset facial expression recognition}. 
	Following the work in~\cite{long2015learning}, we first plug the multi-kernel maximum mean discrepancy (MK-MMD) into the original deep network architecture to effectively measure the divergence between the source and target domain.
	Different from earlier approaches that assume the conditional distributions across domains to be unchanged and learn the invariant representations by matching the marginal distribution, the ECAN minimizes the discrepancy of both marginal and conditional distributions across domains, through making the most use of underlying label information on target data.
	Moreover, we find that the skew class distribution between the source and target domain is a possible bottleneck for cross-domain expression recognition. 
	To address this class distribution bias problem, we introduce a learnable class-wise weighting parameter to the original MMD, in which case, the re-sampled source data can share similar class distribution with the target set. 
	
	Jointly trained with the classical softmax loss which learns distinguishable features under the supervision of the labeled source domain and a variety of MMD regularization terms which improve the generalization ability to the unseen data by exploring the latent semantic meaning behind the unlabeled target domain, the ECAN thus excels in the field of cross-domain FER. 
	The contributions of this paper are as follows:
	
	(1) Our work makes a first step to analyze the facial expression dataset bias quantitatively and explicitly indicates the intrinsic causes of the observed effect. Specifically, our dataset recognition experimental results demonstrate that each dataset tends to have its specific characteristics and preferences during the construction process. And this\textit{ capture bias} will lead to the discrepancy of the marginal distribution. Our cross-dataset generalization experiment further reveals that annotators in each dataset tend to have inconsistent perceptions on the expression categories. And due to this \textit{category bias}, the assumption that the conditional distribution remains unchanged across domains fails to hold in FER applications.
	
	(2) Different from previous deep learning methods, e.g., fine-tuning techniques, that still require a certain number of labels in test set, the unsupervised transfer learning method ECAN is more practical since it is applicable without labeled data in target domain. Moreover, ECAN can learn domain invariant and class discriminative feature representations by effectively aligning the marginal distribution globally and also matching the conditional distributions across domains end-to-end. Furthermore, ECAN takes the expression class distribution bias into account, and embeds a learnable class-wise weighted parameter into the adversarial learning process so that the re-sampled source data and the target data can abide by similar class distribution statistics.

	(3) Extensive experiments have been conducted on different well-established facial expression databases in the literature, including lab-controlled datasets (CK+, JAFFE, MMI and Oulu-CASIA) and real-world databases (AffectNet, FER2013, RAF-DB 2.0 and SFEW 2.0). Comparing with other state-of-the-art methods for cross-dataset facial expression recognition, our ECAN yields comparable and even superior performances.

	This journal paper is an extension of our conference work \cite{li2018deep}.  In the paper, the new contents include the extensive experiments that take a deeper look at expression biases on several widely-used datasets, and expression-conditional extension of domain adaptation method, as well as the corresponding improved performances on various databases.	
	The rest of paper is organized as follows. Section \ref{sec:relate} reviews related work. In Section \ref{sec:look}, we present a deeper look at bias among different facial expression datasets. Section \ref{sec:method} introduces our approach ECAN that learns domain invariant and discriminative features. The results of our experimental evaluation and  experimental analysis are presented in Section \ref{sec:experiment}, and conclusions are drawn in Section \ref{sec:conclusion}.
	
	\section{Related Work}
	\label{sec:relate}
	Numerous approaches of domain adaption have been proposed in the last years to address adaption problems that arise in different computer visual scenarios. 
	However,  few significant interest until now has been gained in the cross-domain learning for facial expression recognition. So besides cross-dataset facial expression recognition, in this section we will also investigate recent works that is closely related to it including cross-domain facial action units detection, micro-expression recognition and pain recognition.
	Generally, we can divide these domain adaption methods into three categories: instance-, model-, and feature-based methods.
	
	Instance based methods re-weight source samples to reduce the distribution differences among the source and target domains.
	In~\cite{miao2012cross}, a supervised extension of Kernel Mean Matching  is proposed to re-weight the source data such that the distributions between source and target domains can be minimized in a class-to-class manner. However, a limited number of labeled data in target domain is required in this case to ensure the discriminative information between classes. 
	Likewise, Chen et al. \cite{chen2013learning} employ the transductive transfer learning algorithm to re-weight each source sample according to the probability ratio between the marginal distributions of the source and the target data, and the target labels is not required in this process.
	In \cite{chu2017selective}, Chu et al.  propose Selective Transfer Machine to re-weight the source training samples that are most relevant to the target users based on an iterative minimization procedure to form a distribution closer to the target data.  However, this instance re-weighting method requires highly expensive computations and is especially time consuming at training time due to the proposed optimization update strategy. On the other hand, user-specific adaptation algorithms are required to be computationally efficient to be used in real world applications. 
	
	Model based methods  construct an adaptive classifier that have parameters or priors shared between the source and the target data. 
	Sangineto et al. \cite{sangineto2014we} propose a personalized model that uses a regression model to learn the mapping from each source subject's data distribution into the associated classifier, then transfers the learned 
	parameters to the unseen 
	target individual. 
	Based on this work, Zen et al. \cite{zen2016learning} further extend the framework to other classification models and use support vectors for parameter transfer. 
	In \cite{eleftheriadis2017gaussian}, a Gaussian process (GP) domain expert is proposed to facilitate the adaption of the classifier by conditioning the target GP on the predictions from multiple source GPs. And the predictions from the individual GP are combined to form the final prediction. 
	However, it still requires a small amount of labeled data of the target subject.
	Huang et al. \cite{huang2018fast} also construct multiple weak generic classifiers based on a subset of the source subjects' data and use the auxiliary target labeled data to evaluate and re-weight each of the weak generic classifiers. And this method also requires annotated target data, which is typically not easily obtained in human computer interaction scenarios.
	Besides, the subject identity information, i.e., the annotation of the person specific data, is required in the above mentioned methods, which is usually not available in the real-world applications.
	
	Feature based methods concentrate on learning features that are domain invariant between source training data and target test data, hence the classifier trained on the source samples can be well fitted into  target data. 
	In~\cite{yan2016transfer}, the author investigates the cross-dataset FER problem and proposes a transfer subspace learning method to obtain a feature subspace that transfers the knowledge gained from the source set to the target data.
	In~\cite{zhu2016discriminative}, a discriminative feature adaptation method is proposed to obtain a feature space that can minimize the mismatch between and source and target distribution.
	In~\cite{yan2016cross}, unsupervised domain adaptive dictionary learning is proposed to learn new representations of original source and target samples that reduce the distribution mismatch between source and target domains.
	Huang et al. \cite{huang2018fast} proposed to align the feature boundaries of different subjects to the same points in the normalized feature space with respect to the neutral expression and the boundary values to remove subject-dependent bias. 
	In \cite{zheng2018cross}, a transductive 
	transfer subspace learning method is proposed to deal with cross-pose and cross-database FER. Specifically, an auxiliary image set from unlabeled target domain is leveraged into the labeled source domain data set for jointly learning a discriminative subspace to reduce the dissimilarity of the marginal probability distributions between the source and target data.
	Most recently, a domain regeneration framework has been proposed in~\cite{zong2018domain} to re-generate source and target samples sharing the same or similar feature distribution for the cross-database micro-expression recognition problem. 
	However, most of these methods neglect the conditional distribution bias between the source and target domains, which is a common phenomenon in realistic conditions. 
	
	
	Since annotating abundant facial expression images from diverse domains
	is a difficult and time consuming work and labeling samples in target domain on-the-fly is impossible, supervised approach that requires labeled instances from the target domain is impractical for real-world applications. So in this paper, we will hinge on unsupervised domain adaption technique for cross-dataset FER. As a few related studies \cite{miao2012cross, chen2013learning, eleftheriadis2017gaussian, huang2018fast} have demonstrated that the supervised information on target data is necessary to boost the adaption performance and enhance the discriminative ability of the learned model, we further explore the underlying semantic meaning behind target data using the pseudo label. And both marginal and conditional distributions will be taken into account to alleviate the domain discrepancy. Moreover, as deep learning has been gradually leveraged to deal with challenging variety in the wild \cite{li2017reliable,li2019Occlu}, we also employ CNN to learn features with more discriminative ability. Hence our approach is unsupervised and can learn discriminative features that generalize well to new domains in which no prior knowledge is available.
	
	To the best of our knowledge, our approach is among the very state-of-the-art ones to  derive feature representations with both generalization and discrimination ability for the cross-domain facial expression recognition problem based on the deep learning technique, where no labeled data are required in the target domain. And for the first time, conditional distribution bias and expression class imbalance problem have been taken into consideration in this case 
	to further boost the final classification performance.
	

	\section{Measuring facial expression dataset bias}
	\label{sec:look}
	Facial expression recognition is a data-driven task. Many databases have been established to evaluate affect recognition systems. However, there exists inevitable bias between these facial expression datasets. 
	Some example face images from different databases have been displayed in Fig. \ref{fig:bias}.
	Previous across-database experiments have also shown that it's hard to learn a classifier which can take into account of all possible variations of the data \cite{COHN2019407}. For example, Shan et. al.~\cite{shan2009facial} trained the Boosted-LBP based SVM classifier on the Cohn–Kanade database, and then evaluated it on the MMI database and the JAFFE database, respectively. The generalization performance across databases has reduced by around 36\% for MMI and 40\% for JAFFE. In~\cite{littlewort2006dynamics}, Littlewort et. al. have also evaluated their systems' generalization ability across databases, which achieved strong performance results within the database (more than 90\%) but decreased to 50--60\% when tested on another different database. Similar observations have been also reported in ~\cite{el2014fully, gu2012facial, whitehill2009toward}. They all demonstrate that dataset bias is very common among different facial expression databases, and evaluating methods with intra-database protocol would render them lack generalization capability on unseen samples at test time.

	Although previous work has demonstrated that dataset bias problem causes a noticeable deterioration in cross-dataset facial expression expression performance, there have been rare discussions in the literature regarding how these facial expression datasets differ among each other. 
	So in this section, we conduct a series of comprehensive experiments including dataset recognition tasks and class-specific cross-dataset generalization tests to evaluate this largely neglected issue quantitatively and widen the attention on concrete aspects in which bias sneaks into these datasets and affects classification performances. Specifically,  a variety of well known lab-controlled \& in-the-wild  facial expression datasets have been evaluated in our experiments, including AffectNet, CK+, FER2013, MMI, Oulu-CASIA, RAF-DB 2.0 and SFEW 2.0.

	\subsection{Dataset Recognition}
	We first evaluate to what extent these facial expression datasets vary across each other, and dig out the main cause of this discrepancy.
	During the image pre-processing stage, facial images from all datasets are first aligned by an affine transformation defined by the centers of the two eyes and the center of the two corners of the mouth and then normalized to the size of $100\times100$.
	Then for image representations, we extract two different features for these already aligned face images: the handcrafted HOG features and the deep learned features. For HOG~\cite{dalal2005histograms} features, we first divide the images into $10\times10$ pixel blocks of four $5\times5$ pixel cells with no overlapping. By setting 10 bins for each histogram, we obtain a 4,000-dimensional feature vector per aligned image. 
	For deep features, we first resize the aligned face images so that the shorter side is 256 pixels, then we crop a $224\times224$ pixels region of each resized image as the network input. We choose the ResNet-50 Convolutional Neural Network \cite{he2016deep} that are first pre-trained on the  MS-Celeb-1M face dataset \cite{guo2016ms} and then fine-tuned on the VGGFace2 face dataset~\cite{cao2018vggface2} to extract deep features. Finally a 2048-dimensional descriptor per image is obtained from the layer adjacent to the classifier layer.
	
\begin{figure}[t]
	\centering
	\subfigure[ Dataset recognition rate]{\label{fig:dr1}
		\includegraphics[width=4cm]{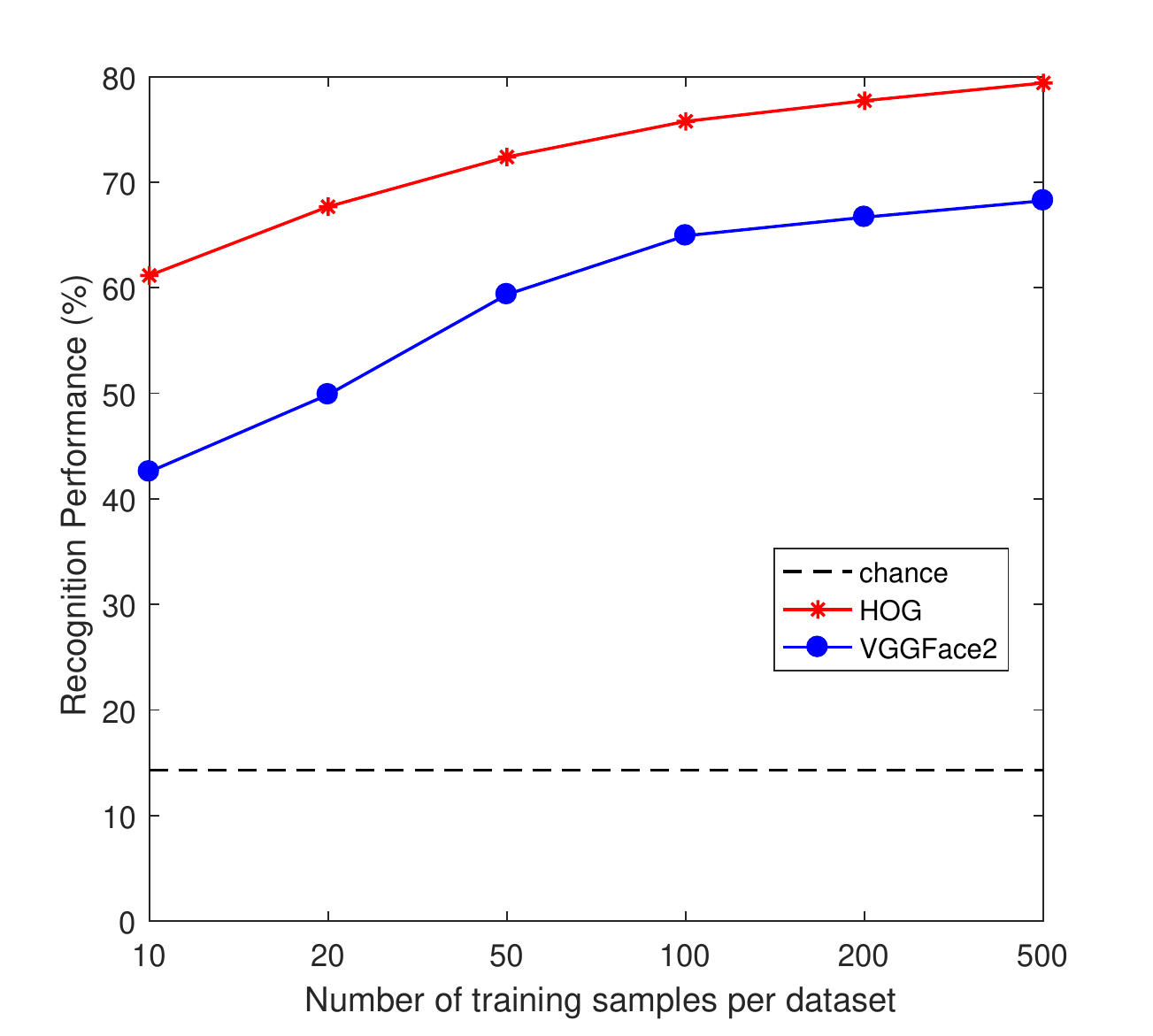}}
	\subfigure[Confusion matrix]{\label{fig:dr2}
		\includegraphics[width=4cm]{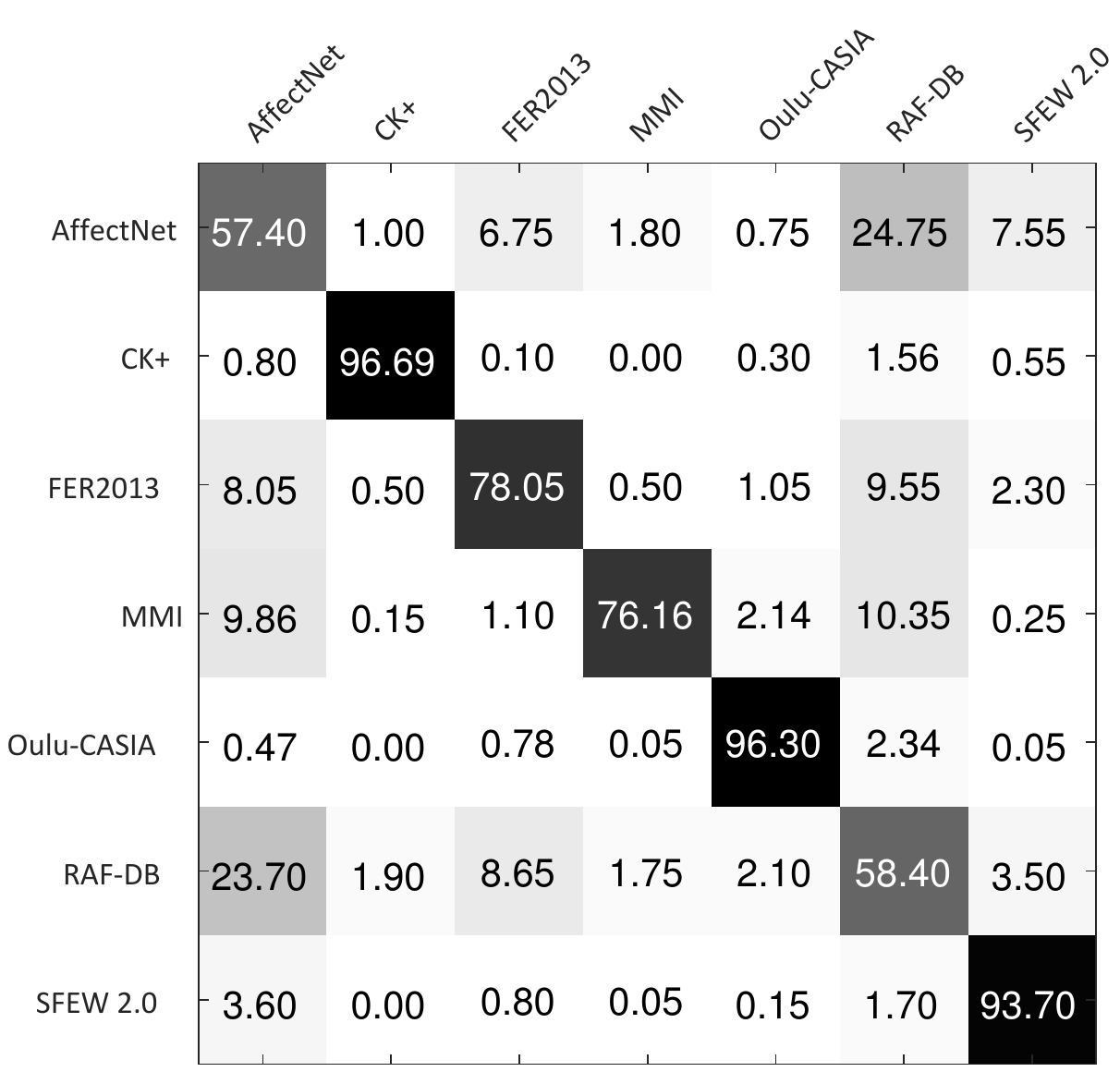}}
	\caption{Dataset recognition experiment on seven different datasets. (a) Classification performance as a function of training set size for two different descriptors. (b) Confusion matrix for the HOG descriptor.}
	\label{fig:dr}
\end{figure}

	To name the dataset, we randomly sample 500 training images from each of the seven datasets and apply a seven-way linear SVM implemented by LibSVM \cite{CC01a}. The penalty parameter $C$ of SVM is set to the default value 1. To ensure person independence, 200 images sampled from each of the remaining sets with  person identities that do not appear in the training set are chosen for classification.
	We repeat it 10 times with different data splits and report the obtained average results. Figure \ref{fig:dr1} shows dataset recognition performance for two feature representations. Figure \ref{fig:dr2} shows the detailed confusion matrix using the HOG feature. 
	
\begingroup
\renewcommand\arraystretch{1.1}
\setlength{\tabcolsep}{3.5pt}
\begin{table*}[t]
	\caption{Cross-dataset generalization on different tasks with two different features. Each matrix contains the expression detection or classification performance when training on one dataset (rows) and testing on another (columns). The diagonal bold elements correspond to the in-dataset results, i.e., training and testing on the same dataset. The ``Mean Others'' refers the average cross-dataset performance on all the other test datasets except self. For short, A = AffectNet, C = CK+, F = FER2013, M = MMI, O = Oulu-CASIA, R = RAF-DB 2.0, S = SFEW 2.0.}
	\label{tab:cross}
	\noindent\begin{tabular}{|c|m{6.8cm}<{\centering} c p{0.6cm}<{\centering} |m{6.8cm}<{\centering} c p{0.6cm}<{\centering}|}
		\hline 
		&HOG & \begin{tabular}[c]{@{}c@{}} \scriptsize Mean\\ \scriptsize Others\end{tabular} & \begin{tabular}[c]{@{}c@{}} \scriptsize Percent \\\scriptsize Drop \end{tabular}&Deep feature & \begin{tabular}[c]{@{}c@{}} \scriptsize Mean\\ \scriptsize Others\end{tabular} & \begin{tabular}[c]{@{}c@{}} \scriptsize Percent \\\scriptsize Drop \end{tabular} \\
		\hline  & & &&&& \\[-2.5ex]
		\multirow{8}{*}{\rotatebox{90}{Happiness}} &
		\noindent\multirow{8}{*}{
			\begin{tabular}{|>{\scriptsize}c|*{7}{>{\scriptsize}p{0.5cm}<{\centering}|}}
				\hline
				\cellcolor{gray!20}\diagbox[height=2em,trim=l]{Train}{Test}&\cellcolor{gray!20}A & \cellcolor{gray!20}C& \cellcolor{gray!20}F &\cellcolor{gray!20}M & \cellcolor{gray!20}O &\cellcolor{gray!20}R &\cellcolor{gray!20}S \\ \hline
				\cellcolor{gray!20}\tiny AffectNet & \textbf{89.14} & 95.55 & 80.41 &91.90 &90.42 &75.38&84.64  \\ \hline
				\cellcolor{gray!20}\tiny CK+ &88.69 & \textbf{98.96} & 79.53 & 89.77 & 92.40 & 72.06 & 83.23  \\ \hline
				\cellcolor{gray!20}\tiny FER2013 & 86.23 & 81.63 & \textbf{87.05} & 90.91 & 90.68 & 82.40&77.82 \\ \hline
				\cellcolor{gray!20}\tiny MMI & 88.80 & 97.01 & 81.74 & \textbf{94.65}
				& 94.90 & 76.71 &83.01\\ \hline
				\cellcolor{gray!20}\tiny Oulu-CASIA & 86.60 & 90.29 & 77.71 & 87.50 & \textbf{96.98} & 70.37 &81.01\\ \hline
				\cellcolor{gray!20}\tiny RAF-DB 2.0& 83.40 & 91.02 & 82.03 & 91.62 & 86.62 & \textbf{83.45}&78.26 \\ \hline
				\cellcolor{gray!20}\tiny SFEW 2.0& 87.23 & 91.58 & 80.05 & 90.62 & 89.69 & 75.55 &\textbf{88.80}\\ \hline
			\end{tabular}
		} & & &
		\noindent\multirow{8}{*}{
			\begin{tabular}{|>{\scriptsize}c|*{7}{>{\scriptsize}p{0.5cm}<{\centering}|}}
				\hline
				\cellcolor{gray!20}\diagbox[height=2em,trim=l]{Train}{Test}&\cellcolor{gray!20}A & \cellcolor{gray!20}C& \cellcolor{gray!20}F &\cellcolor{gray!20}M & \cellcolor{gray!20}O &\cellcolor{gray!20}R &\cellcolor{gray!20}S \\ \hline
				\cellcolor{gray!20}\tiny AffectNet & \textbf{89.51} & 93.61  & 87.03 &92.33 &93.59&81.77&84.64  \\ \hline
				\cellcolor{gray!20}\tiny CK+ & 85.23 & \textbf{94.18} & 84.82 & 91.19 & 93.02 & 79.97 &81.68 \\ \hline
				\cellcolor{gray!20}\tiny FER2013 & 90.51 & 94.90 & \textbf{91.56} & 90.34 & 92.03 & 84.64&87.24 \\ \hline
				\cellcolor{gray!20}\tiny MMI & 83.77 & 90.94 & 82.44 & \textbf{88.20} & 90.78 & 77.78 &78.56\\ \hline
				\cellcolor{gray!20}\tiny Oulu-CASIA & 86.34 & 91.67 & 82.89 & 92.90 &\textbf{94.64}& 76.67 &78.26\\ \hline
				\cellcolor{gray!20}\tiny RAF-DB 2.0& 82.71 & 89.48 & 82.79 & 84.38 & 81.93 &\textbf{90.15}&82.20 \\ \hline
				\cellcolor{gray!20}\tiny SFEW 2.0& 84.57 & 87.54& 81.80 & 87.22 & 86.15 & 76.33 &\textbf{89.17}\\ \hline
			\end{tabular}
		} &  \\ [0.5ex]
		&&86.38&3\% & &88.83&1\%\\ [0.3ex]
		&&84.28&15\%&&85.99&9\%\\[0ex]
		&&84.94&2\%&&89.95&2\%\\[0.ex]
		&&87.03&8\%&&84.05&5\%\\[0.3ex]
		&&82.25&15\%&&84.79&10\%\\[0.2ex]
		&&85.49&-2\%&&83.91&7\%\\[0.2ex]
		&&85.79&3\%&&83.93&6\%\\[0.5ex]
		\hline  & &&&&& \\[-2.5ex]
		\multirow{8}{*}{\rotatebox{90}{Sadness}} &
		\noindent\multirow{8}{*}{
			\begin{tabular}{|>{\scriptsize}c|*{7}{>{\scriptsize}p{0.5cm}<{\centering}|}}
				\hline
				\cellcolor{gray!20}\diagbox[height=2em,trim=l]{Train}{Test}&\cellcolor{gray!20}A & \cellcolor{gray!20}C& \cellcolor{gray!20}F &\cellcolor{gray!20}M & \cellcolor{gray!20}O &\cellcolor{gray!20}R &\cellcolor{gray!20}S \\ \hline
				\cellcolor{gray!20}\tiny AffectNet & \textbf{80.97} & 86.97 & 77.61 &85.65 &73.02 &82.24&76.33  \\ \hline
				\cellcolor{gray!20}\tiny CK+ & 84.97 & \textbf{96.63}
				& 84.08 & 88.21 & 87.92 & 85.99 &81.53 \\ \hline
				\cellcolor{gray!20}\tiny FER2013 & 80.29 & 89.89 & \textbf{81.96} & 85.08 & 80.10 & 82.79&71.74 \\ \hline
				\cellcolor{gray!20}\tiny MMI & 84.00 & 93.45 & 83.11 & \textbf{89.71} & 85.62 & 85.24 &78.71\\ \hline
				\cellcolor{gray!20}\tiny Oulu-CASIA & 79.83 & 91.99 & 77.57 & 83.52 & \textbf{91.20} & 80.78 &79.01\\ \hline
				\cellcolor{gray!20}\tiny RAF-DB 2.0& 80.89 & 89.00 & 77.83 & 78.27 & 80.99 & \textbf{83.02}&79.67 \\ \hline
				\cellcolor{gray!20}\tiny SFEW 2.0& 73.86 & 86.41 & 71.51 & 84.80 & 76.15 &74.12 &\textbf{83.98}\\ \hline
			\end{tabular}
		} & & &
		\noindent\multirow{8}{*}{
			\begin{tabular}{|>{\scriptsize}c|*{7}{>{\scriptsize}p{0.5cm}<{\centering}|}}
				\hline
				\cellcolor{gray!20}\diagbox[height=2em,trim=l]{Train}{Test}&\cellcolor{gray!20}A & \cellcolor{gray!20}C& \cellcolor{gray!20}F &\cellcolor{gray!20}M & \cellcolor{gray!20}O &\cellcolor{gray!20}R &\cellcolor{gray!20}S \\ \hline
				\cellcolor{gray!20}\tiny AffectNet & \textbf{81.23}& 81.88  & 79.77 &78.27 &74.37 &78.51&76.11 \\ \hline
				\cellcolor{gray!20}\tiny CK+ & 83.06 & \textbf{94.18} & 82.67 & 88.35 & 87.50 & 84.07 &80.93 \\ \hline
				\cellcolor{gray!20}\tiny FER2013 & 83.63 & 91.02 & \textbf{85.19} & 87.07 & 87.19 &86.72&78.19 \\ \hline
				\cellcolor{gray!20}\tiny MMI & 80.66 & 89.81 & 79.85 & \textbf{85.15} & 82.45 & 80.98 &76.93\\ \hline
				\cellcolor{gray!20}\tiny Oulu-CASIA & 68.66 & 79.37 & 71.88 & 76.85 & \textbf{82.08} & 69.37 &73.89\\ \hline
				\cellcolor{gray!20}\tiny RAF-DB 2.0& 85.86 & 91.10 & 84.32 & 87.36 & 86.35 & \textbf{89.52}&81.08 \\ \hline
				\cellcolor{gray!20}\tiny SFEW 2.0&79.23& 87.22 &79.10 & 76.85 & 79.58 & 80.05 & \textbf{84.64} \\ \hline
			\end{tabular}
		} & \\ [0.5ex]
		&&80.31&1\% & &78.15&4\% \\ [0.3ex]
		&&85.45&12\%&&84.43&10\%\\[0ex]
		&&81.65&0\% &&85.64&-1\% \\[0.ex]
		&&85.02&5\%&&81.78&4\%\\[0.3ex]
		&&82.12&10\%&&73.34&11\%\\[0.2ex]
		&&81.11&2\%&&86.01&4\%\\[0.2ex]
		&&77.81&7\%&&80.34&5\%\\[0.5ex]
		\hline  & & &&&& \\[-2.5ex]
		\multirow{8}{*}{\rotatebox{90}{7 expressions}} &
		\noindent\multirow{8}{*}{
			\begin{tabular}{|>{\scriptsize}c|*{7}{>{\scriptsize}p{0.5cm}<{\centering}|}}
				\hline
				\cellcolor{gray!20}\diagbox[height=2em,trim=l]{Train}{Test}&\cellcolor{gray!20}A & \cellcolor{gray!20}C& \cellcolor{gray!20}F &\cellcolor{gray!20}M & \cellcolor{gray!20}O &\cellcolor{gray!20}R &\cellcolor{gray!20}S \\ \hline
				\cellcolor{gray!20}\tiny AffectNet & \textbf{48.83} & 58.58  & 28.29 &42.19 &32.50 &33.33&27.74 \\ \hline
				\cellcolor{gray!20}\tiny CK+ & 34.09 & \textbf{93.92} & 31.07 & 56.68 & 58.85 & 34.96 &25.59 \\ \hline
				\cellcolor{gray!20}\tiny FER2013 & 30.14 & 44.58 & \textbf{52.28} & 45.60 & 41.30 & 46.44&33.98 \\ \hline
				\cellcolor{gray!20}\tiny MMI & 31.83 & 74.11 & 32.73 & \textbf{65.11} & 53.85 & 36.34 &26.71\\ \hline
				\cellcolor{gray!20}\tiny Oulu-CASIA & 24.69 & 74.68 & 27.01 & 47.30 & \textbf{73.59} & 25.80 &18.55\\ \hline
				\cellcolor{gray!20}\tiny RAF-DB 2.0& 32.71 & 59.87 & 39.05 &51.70 &44.84
				& \textbf{55.36} &30.49\\ \hline
				\cellcolor{gray!20}\tiny SFEW 2.0& 26.17 & 40.86 & 30.94 & 35.80 & 31.87 & 33.33 &\textbf{55.05}\\ \hline
			\end{tabular}
		} & & & 
		\noindent\multirow{8}{*}{
			\begin{tabular}{|>{\scriptsize}c|*{7}{>{\scriptsize}p{0.5cm}<{\centering}|}}
				\hline
				\cellcolor{gray!20}\diagbox[height=2em,trim=l]{Train}{Test}&\cellcolor{gray!20}A & \cellcolor{gray!20}C& \cellcolor{gray!20}F &\cellcolor{gray!20}M & \cellcolor{gray!20}O &\cellcolor{gray!20}R &\cellcolor{gray!20}S \\ \hline
				\cellcolor{gray!20}\tiny AffectNet &\textbf{58.43}& 51.38& 39.00& 45.60 &37.76&42.30 &29.67  \\ \hline
				\cellcolor{gray!20}\tiny CK+ & 32.28 & \textbf{85.40} & 41.75 & 52.84 & 53.49 & 40.76&30.41 \\ \hline
				\cellcolor{gray!20}\tiny FER2013 & 36.54 & 58.25 & \textbf{61.74} & 46.87 & 43.12 & 52.70&34.72 \\ \hline
				\cellcolor{gray!20}\tiny MMI & 29.11 & 51.94 & 32.08 & \textbf{53.19} & 40.68 & 32.80 &24.03\\ \hline
				\cellcolor{gray!20}\tiny Oulu-CASIA & 25.57 & 58.58 & 30.96 & 47.02 & \textbf{61.25} & 30.16 &21.37\\ \hline
				\cellcolor{gray!20}\tiny RAF-DB 2.0& 38.29 & 58.09 & 49.87 & 48.44 & 45.52 & \textbf{68.69}&38.13 \\ \hline
				\cellcolor{gray!20}\tiny SFEW 2.0& 27.23 & 36.00 & 33.67 & 23.15 & 29.58 & 33.22 &\textbf{58.09}\\ \hline
			\end{tabular}
		} &  \\ [0.5ex]
		&&36.92&24\%& &40.95&30\% \\ [0.3ex]
		&&40.21&57\%&&41.92&51\%\\[0ex]
		&&40.34&23\%&&45.37&27\% \\[0.ex]
		&&42.59&35\%&&35.11&34\%\\[0.3ex]
		&&36.34&51\%&&35.61&42\%\\[0.2ex]
		&&43.11&22\%&&46.39&32\%\\[0.2ex]
		&&33.16&40\%&&30.48&48\%\\[0.5ex]
		\hline
	\end{tabular}
\end{table*}
\endgroup
	From the plot on the left, we can see that the best classification performance reaches 79\% (the chance is 1/7=14\%) despite the small sample size, and there is no evident tendency of saturation as more training data is added. When comparing the performances of two different features, it indicates that HOG feature shows an advantage in separating these datasets than deep feature. As the deep network is pre-trained on large-scale face datasets, features extracted from it are more generalized and universal. Hence deep learning can learn common representations and achieve higher generalizability across domains.
	From the confusion matrix of HOG feature on the right, we see that it is easy to distinguish lab-controlled datasets, especially CK+ and Oulu-CASIA datasets, from all the others. And there is a confusion grouping among real-world datasets. Furthermore, AffectNet and RAF-DB 2.0 are the datasets with the highest confusion degree, which is mainly due to the diverse and extensive image samples they contain.
	
	In summary, there appears to have obvious distinctions among these expression datasets. And the main consequences of this discrepancy can be accounted for the capture bias, that is, each dataset tends to have its own preference during the construction processing. 
	For example, images can be acquired with different devices (e.g., professional pictures photographed using the laboratory camera \textit{vs.} amateur snapshots and selfies collected from the Internet), and various collection environments (e.g., different lighting conditions and occlusions, certain types of background, post-processing elaboration such as filtering). In addition, participants' identity and character in these datasets are also different. For instance, persons in MMI dataset carry more accessories, e.g., glasses and mustache, when compared to other lab-controlled datasets. And datasets retrieved from the Internet contain subjects with a much wider range of age and ethnicity.
	
	The capture bias makes these datasets following different marginal distributions, which is the public acknowledged issue to be solved in recent studies on domain adaption. And in the following subsection, we will look into another generally ignored but equally important issue, i.e., the conditional distribution of each dataset.
	
	\subsection{Cross-dataset Generalization}
	\label{cross}
	In this experiment, we investigate how well does an expression class model trained on one dataset generalize when tested on other datasets, and give an insight on how the expression classes in each dataset are associated with each other. 
	
	We study the class-specific cross-dataset performances on these datasets with two testing regimes: \textit{facial expression detection} that detects images with one specific expression class from all images; and \textit{facial expression classification} that finds all images containing the desired expression class. Notice that the classification task is basically the same as the detection task if there are only two classes. And we follow the image representations (HOG and deep feature) and classifier (linear SVM) used in the above data recognition experiment.
	
	For the \textit{expression detection task}, we use all images with the target expression label in each dataset as the positive samples and the others as the negative samples. In the in-database setting, we follow the person-independent rule and conduct five-fold cross-validation experiment on each dataset. In the cross-database setting, we train our model using all images in one dataset and  test it on another dataset. And we focus on two expressions that are representative and common in all the datasets: happiness and sadness. Likewise, for the \textit{expression classification task}, we also adopt all images in each dataset and follow the person-independent protocol.
	Table \ref{tab:cross} present results for each task. Note that since each dataset contains different number of images per class and has different difficulty degree on expression recognition, the actual accuracy values for each dataset do not mean a lot to the generalization performance. And it is the percent drop between the in-dataset performance and  the average cross-dataset performance over other test datasets that bears much more important information on the evaluation for cross-dataset generalization.
	
	First, there is a significant drop of performance when testing on other different datasets,  on the whole, in all tasks and features. For instance, the average in-dataset performance of all datasets is 63.83\% in the seven expression classification task using deep feature, and it drops to 39.40\% for the average cross-dataset performance. This observation is coherent with what we deduce from the data recognition experiment, which verifies the distinct bias across these datasets again. Second, comparing different features on the in-database performance, the deep feature shows the superiority on real-world datasets, which mainly benefits from the large-scale in-the-wild dataset the network pre-trained on. However, as a byproduct trained on face recognition dataset, the deep feature performs worse on the lab-controlled dataset with person-independent setting. And for the cross-database performance, deep feature achieves better results on the average mean other accuracy and percent drop, which indicates that the deep feature captures generalized information and is more suitable for domain adaption. Third, focusing on the difficulty and diversity of each dataset, lab-controlled datasets, especially CK+, are easier across all tasks (column average). And real-world datasets, especially AffectNet and RAF-DB 2.0, are more diverse and generalized (row average), which even achieve higher cross-dataset performance on lab-controlled datasets than the in-dataset one for several tasks. 
	
	
	An important factor that intrinsically induces this declination on class-specific cross-dataset generalization is the category bias: annotators in each dataset, including human and machine, may have different interpretations and perceptions of the emotion conveyed in facial images, and many images tend to express more than one expression category which further enhances the difficulty and uncertainty of annotation. Due to this fact, the conditional distributions in these datasets are indeed different. However, many researches assume that the conditional probability distributions remain the same across different datasets before conducting domain adaption, which would prevent good domain invariant feature representations from being learned.
	

	\section{Emotion-Conditional Adaption Network}
	\label{sec:method}
	Although the above experiments demonstrate the generalization ability of deep feature, the  dataset bias still remains evident. So in this section, we will focus on this problem and present our approach for dealing with cross-dataset facial expression recognition. For the convenience of reading, we list some of the major notations that are used throughout this paper in Table \ref{tab:notation}.

	\subsection{Problem Formulation}
	We first start from the problem definition. Let $\mathcal{\mathbf{X}}_s\in \mathbb{R}^{D\times N_s}$ be the source domain data with marginal probability distribution $P^s(\mathbf{X})$ and $\mathcal{\mathbf{X}}_t\in \mathbb{R}^{D\times N_t}$ be the target domain with marginal probability distribution $P^t(\mathbf{X})$, where $D$ is the dimension of the data instance, $N_s$ and $N_t$ are number of images in source and target domain respectively. 
	
	In unsupervised domain adaption, we refer to the training dataset with labeled data as the source domain  \(\mathcal{D}_s=\{(\bm{x}_i^s, y_i^s)\}_{i=1}^{N_s}\) where $y_i^s\in \{1,2,3,4,5,6,7\}$ is the corresponding expression class label of source data $\bm{x}_i^s$, and the test dataset without labeled data as the target domain $\mathcal{D}_t=\{(\bm{x}_i^t)\}_{i=1}^{N_t}$. 
	Both the source and target data pertain to seven expression classes in this case. 
	And in this scenario, the source and target set are assumed to be different but related due to different acquisition conditions. Our goal is to learn a deep neural network $f:\mathcal{X}\rightarrow\mathcal{Y}$ that can decrease the cross-domain discrepancy so as to minimize the target error just using the supervision information from source data.
		\begin{table}[t]
		\caption{Summary of Major Notations Used in the Paper} 
		\small
		\renewcommand\arraystretch{1.2}
		\label{tab:notation}%
		\begin{tabular}{ll}
			\toprule
			Notations & Description\\
			\midrule
			$\mathcal{D}_s/\mathcal{D}_t$&the source/target domain\\
			$\bm{x}_i^s/\bm{x}_i^t$ & feature vector of sample in the source/target domain \\
			$N_s/N_t$&number of source/target samples\\
			$y_i^s$ & expression label of source sample $\bm{x}_i^s$ \\
			$\phi(\cdot)$ & feature map function with kernel map $k(\bm{x},\cdot)$\\
			$P(\mathbf{X})$& marginal probability distribution of domain data $\bm{X}$\\
			$P(Y|\bm{X})$&conditional probability distribution of $Y$ given $\bm{X}$ \\
			$\alpha_l$ & weight ratio to sample with label $l$\\
			$\bm{x}_{i,l}^s$&feature vector of source data with true label $l$\\
			$\bm{x}_{i,\hat{l}}^t$&feature vector of target data with pseudo label $\hat{l}$\\
			$N_s^l$& number of source samples with true label $l$\\
			$N_t^{\hat{l}}$&number of target samples with pseudo label $\hat{l}$\\
			$\Theta$&network parameters to be learned\\
			$\gamma,\lambda$&trade-off hyper-parameters\\
			\bottomrule
		\end{tabular}
	\end{table}

	\subsection{Preliminary and Motivation}
	We first introduce Maximum Mean Discrepancy (MMD) which is a widely-used and efficient non-parametric metric on the embeddings of probability distributions to measure the divergence between different domains~\cite{borgwardt2006integrating}. The MMD and its empirical estimate in the reproducing kernel Hilbert space (RKHS) can be defined as:
	\begin{equation}
	\text{MMD}[\mathcal{F},\mathcal{P}_s,\mathcal{P}_t] = \mathop{\text{sup}}\limits_{f\in\mathcal{F}}(\mathbf{E}_{p_s}[f(\bm{x}^s)]-\mathbf{E}_{p_t}[f(\bm{x}^t)]),  \label{eq1}
	\end{equation}
	where $\mathbf{E}_{p_s}$and $\mathbf{E}_{p_t}$ denote the population expectations under distribution $p_s$ and $p_t$, respectively.  The MMD function class $\mathcal{F}$ is the unit ball in a reproducing kernel Hilbert space $\mathcal{H}$. We then map the data into $\mathcal{H}$ using feature space mapping function $\phi(\cdot)$ and get an empirical estimate of the MMD:
	\begin{equation}
	\text{MMD}^2[\mathcal{H},\mathcal{D}_s,\mathcal{D}_t] =\left |\left |\frac{1}{N_s} \sum_{i=1}^{N_s}\phi (\bm{x}_i^s)-\frac{1}{N_t} \sum_{i=1}^{N_t}\phi (\bm{x}_i^t) \right |\right |_\mathcal{H}^2, \label{eq2}
	\end{equation}
	where $\phi(\cdot)$ denotes the feature map function associated with the kernel map $k(\bm{x}^s,\bm{x}^t)=\left \langle\phi(\bm{x}^s),\phi(\bm{x}^t)\right \rangle_{\mathcal{H}}$. One of the most used kernel corresponding to an infinite-dimensional $\mathcal{H}$ is the Gaussian kernel $k(\bm{x}^s,\bm{x}^t)=exp(-||\bm{x}^s-\bm{x}^t||^2/(2\sigma^2))$. 
	To maximize the test power and minimize the test error, the kernel map $\mathcal{K}$ is defined as a linear combination of  base kernels $\{k_u\}_{u=1}^d$:
	\begin{equation}
	\mathcal{K}=\left\{k=\sum_{u=1}^d\beta_uk_u,  \sum_{u=1}^d\beta_u=1, \beta_u\geqslant0, \forall u\right\}. \label{eq3}
	\end{equation}
	Given the statistical tests defined in MMD, we can have $\mathcal{P}_s=\mathcal{P}_t$  if and only if $\text{MMD}=0$.
	Hence we can use the MMD function based on the feature map $\phi$ in Eq. \ref{eq2} to detect any marginal distribution discrepancy between $\mathcal{D}_s$ and $\mathcal{D}_t$.
	To adapt the MMD function to the stochastic gradient descent (SGD) in CNN-based domain adaption, we further adopt an unbiased  empirical estimator \cite{gretton2012kernel} of $\text{MMD}^2[\mathcal{H},\mathcal{D}_s,\mathcal{D}_t]$  with linear time complexity for efficiency:
	\begin{equation}
	\begin{split}
	\text{MMD}_u^2[\mathcal{H},\mathcal{D}_s,\mathcal{D}_t]&=\frac{1}{N_s(N_s-1)}\sum_{i\neq j}^{N_s}k(\bm{x}_i^s,\bm{x}_j^s)\\
	&+\frac{1}{N_t(N_t-1)}\sum_{i\neq j}^{N_t}k(\bm{x}_i^t,\bm{x}_j^t)\\
	&-\frac{2}{N_s N_t}\sum_{i=1}^{N_s}\sum_{i=1}^{N_t}k(\bm{x}_i^s,\bm{x}_j^t).
	\label{eq4}
	\end{split}
	\end{equation}

	Actually, many domain adaption approaches have employed MMD to reduce the marginal distribution distance between source and target domains. However, additional important issues have been ignored:
	
	(1)\textit{ Conditional distribution discrepancy}: As has been mentioned in Sec. \ref{cross}, the conditional distributions in facial expression datasets are indeed different. Therefore, we propose to match not only the marginal distributions but also the conditional distributions across domains by optimizing their empirical MMD distance, respectively.
	
	(2) \textit{Class discriminative capacity}: Domain invariance can effectively transfer knowledge from the source data to the target data. However, it can not guarantee the discriminability of learned features. For example, features with fear label in the target domain can be mapped near features with surprise label in the source domain while also satisfying the condition of global domain-invariance.
	
	(3) \textit{Class imbalanced problem}: the biased distribution is quite frequent in facial expression datasets and exists in most realistic settings. This can be justified by the practicalities of data acquisition: eliciting and annotating a smile is very easy, however, capturing  information for disgust, anger or less common expressions can be very challenging. Samples in one category dominate but lack in another can mislead judgment on the source domain's relevance on the target task. This phenomenon that has been neglected by most previous research would be a significant bottleneck for cross-dataset facial expression analysis. 
	
	So in the following paragraph, we will present our approach to deal with these outstanding issues.
	
	\begin{figure*}
		\centering
		\begin{minipage}{8cm}
			\centerline{\includegraphics[width=7cm]{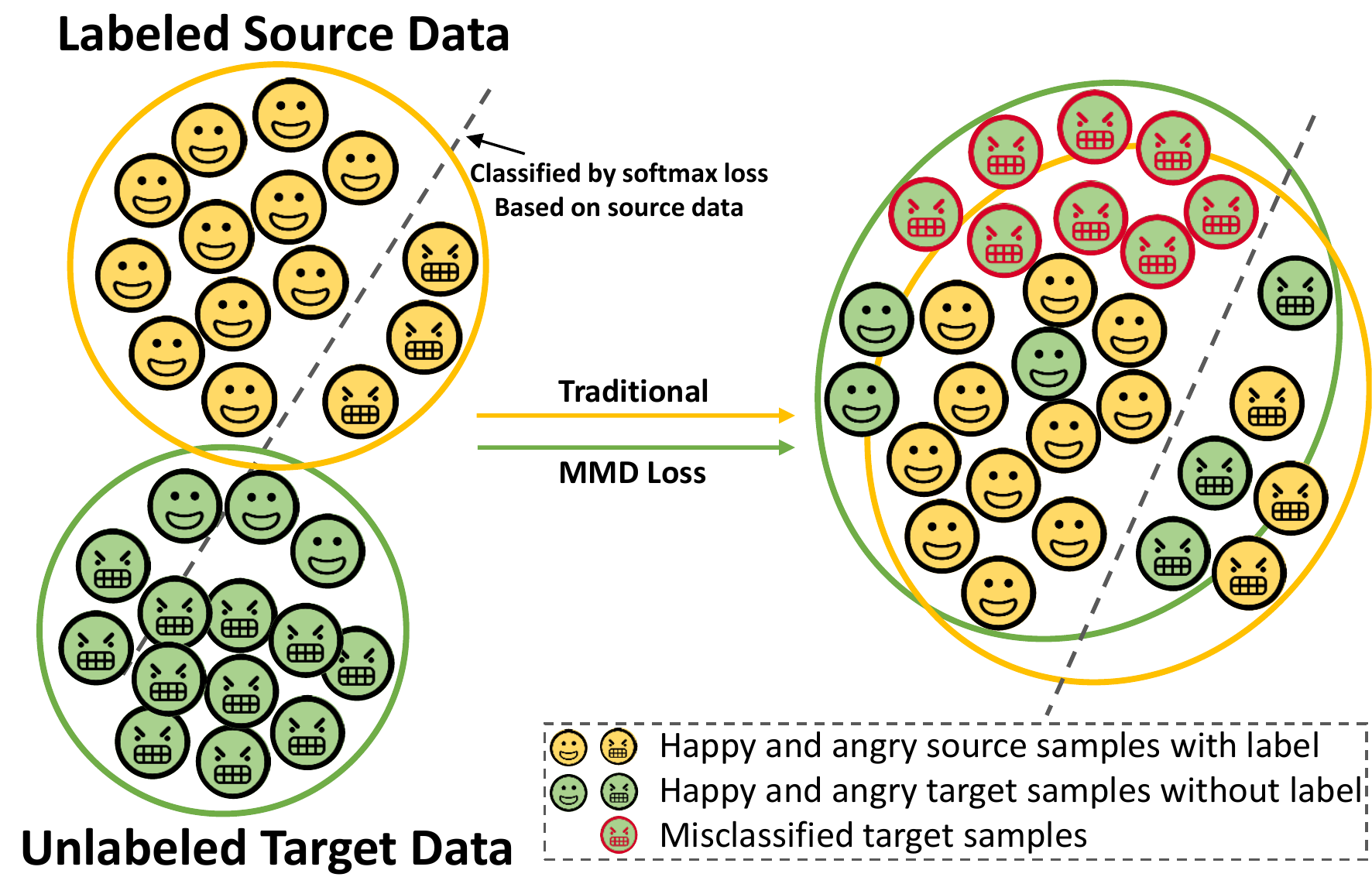}}
			\centerline{(a) Original MMD}  
		\end{minipage}
		\begin{minipage}{10cm}
			\centerline{\includegraphics[width=10cm]{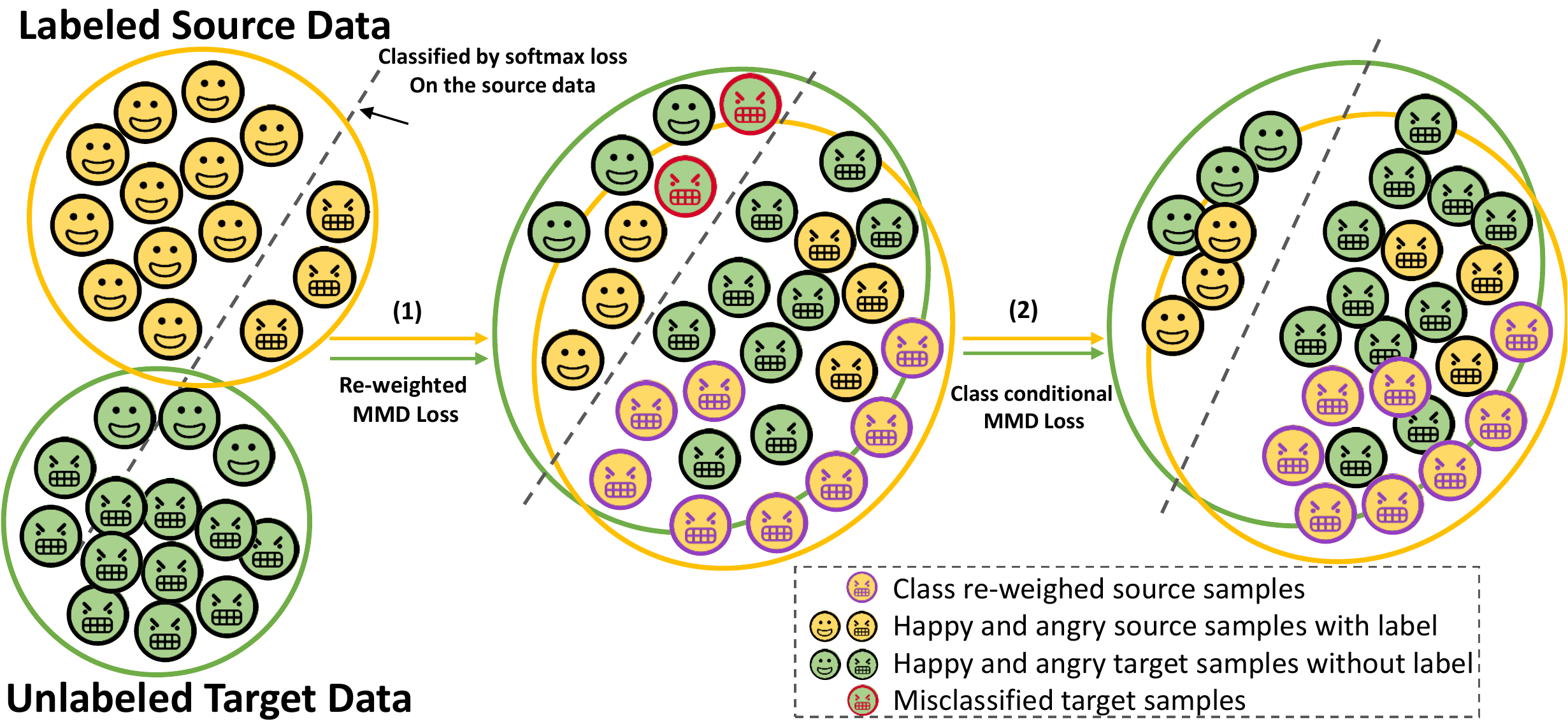}}
			\centerline{(b) ECAN}  
		\end{minipage}
		\caption{(a) In traditional MMD, only the discrepancy of  marginal distributions between source and target domains
		 are considered to be restricted. Since domain-invariance does not mean discriminativeness and class distribution bias exists across domains, samples in target domains are still prone to be misclassified even if the marginal distribution is invariant. As the graph illustrates, the most part of target domain are misclassified as ``happy'' which are dominate in the source domain. (b) In our method, we explore the underlying label information of target data, and 
		 match both the marginal and class conditional distributions to mitigate the discrepancy across domains. With the Re-weighed MMD redistributing the class distribution of the source domain and the class conditional MMD  learning the conditional invariant transformation, the discriminative separating hyperplane thus can generalize well on the target data.}
		\label{fig:method}
	\end{figure*}

	\subsection{Emotion-Conditional Adaption Network (ECAN)}
	\label{method}
	The existing methods propose to match the marginal distribution $P(\bm{X})$ across different domains. However, it cannot guarantee that the discrepancy in the conditional distribution $P(Y|\bm{X})$  is also decreased. Our goal is to learn a deep network that can ensure the invariant of both the marginal and the conditional distribution so that the efficiency of domain adaption across different facial expression datasets can be improved.
	
	As the label variable $Y$ is in a low-dimensional and discrete space, it is more efficient to match label distributions $P(\bm{Y})$ compared with its conditional variant. According to the Bayes rule $P(\bm{X}|Y)P(Y)\propto P(Y|\bm{X})$, we can then correct the changes in conditional distribution by matching $P(\bm{X}^s|Y^s)P(Y^s)$ and $P(\bm{X}^t|Y^t)P(Y^t)$. And for the probability distribution of $\bm{X}$ given $Y$, we propose to learn a conditional invariant feature transformation such that $P(\bm{X}^s|Y^s)=P(\bm{X}^t|Y^t)$. Considering that there is no labeled data in the target domain for unsupervised domain adaption, we further resort to the pseudo label of the target data to explore accessible semantic information of the distributions. We then present this two additional regularization terms that can be intuitively implemented in the deep learning framework. 
	
	(1) \textit{Learning Class Re-weighted Label Space}:
	As has been mentioned before, class imbalance is very common in facial expression databases, especially for the real-world scenarios. During experiments, we found that, in some cases, simply using MMD for cross-domain facial expression recognition would even degrade performances. To remedy this problem, we propose to apply a class re-weighted term to the MMD distance of distribution. 
	Note that the marginal probability distribution of both domains can be written as:
	\begin{equation}
	P(\bm{X})=\sum_{l=1}^7P(\bm{X},Y)=\sum_{l=1}^7P(\bm{X}|Y=l)P(Y=l),
	\label{eq6}
	\end{equation}
	where $P(\bm{X}|Y=l)$ is the conditional distribution of X given Y and $P(Y=l)$ is the class prior probability.
	And the assumption that the class distribution, i.e., the class prior probability, is the same in both domains is no longer valid in our case. When $P(Y)$ is different across domains, $P(\bm{X})$ can still vary even if $P(\bm{X}|Y)$ remains unchanged. In this case, commonly restricting the discrepancy of $P(\bm{X})$ may lead to the effect of misleading information, that is, predictions on certain classes would be easily misguided by differences between label occurrences in both domains. A special example can intuitively explain this phenomenon: Assuming that samples with label fear only  exist in the source domain, if we strictly require the target domain to have the same distribution with the source domain, then  part of samples in the target domain will be incorrectly classified as fear under the supervision of the original MMD algorithm. 
	
	In order to weaken the impact of the class distribution bias, we add a weight ratio $\alpha$ to re-sample the class distribution in the source domain. Then the re-weighted marginal distribution of the source domain can be formulated as:
	\begin{equation}
	\begin{split}
	P_{\alpha}(\bm{x}^s)&=\sum_{l=1}^7P(\bm{X}^s|Y^s=l)P(Y^t=\hat{l})\\
	&=\sum_{l=1}^7P(\bm{X}^s|Y^s=l)P(Y^s=l)\alpha_l
	\label{eq7}
	\end{split}
	\end{equation}
	where the weight ratio $\alpha_l=\frac{P(Y^t=\hat{l})}{P(Y^s=l)}$. 
	For the target domain we do not know the true labels, we use the pseudo labels $\hat{l}$ 
	predicted by the classifier trained on the source domain.
	By amending the original MMD with the weighted ratio $\alpha$, the source domain thus can share the same class distribution with the target domain. 
	Combining Eq. \ref{eq2}, Eq. \ref{eq4} and Eq. \ref{eq7}, the weighted empirical estimate MMD  of two domains and its unbiased estimate can be formulated as:
	
	\begin{equation}
	\text{MMD}_\alpha^2[\mathcal{D}_s,\mathcal{D}_t]=\left |\left |\frac{1}{N_s} \sum_{i=1}^{N_s}\alpha_{y_i^s}\phi (\bm{x}_i^s)-\frac{1}{N_t} \sum_{i=1}^{N_t}\phi (\bm{x}_i^t) \right |\right |^2 
	\end{equation}
	\begin{equation}
	\begin{split}
	\text{MMD}_{\alpha,u}^2[\mathcal{D}_s,\mathcal{D}_t]&=\frac{1}{N_s(N_s-1)}\sum_{i\neq j}^{N_s}\alpha_{y_i^s}\alpha_{y_j^s}k(\bm{x}_i^s,\bm{x}_j^s)\\
	&+\frac{1}{N_t(N_t-1)}\sum_{i\neq j}^{N_t}k(\bm{x}_i^t,\bm{x}_j^t)\\
	&-\frac{2}{N_s N_t}\sum_{i,j=1}^{N_s,N_t}\alpha_{y_i^s}k(\bm{x}_i^s,\bm{x}_j^t)
	\end{split}
	\end{equation}

	(2) \textit{Learning Class-Conditional Invariant Features}: On condition that the invariance of the class prior probability is guaranteed, we propose to learn the conditional invariant transformation by  applying the MMD in the class level. With the true labels in source domain and the pseudo labels in target domain, we can then  match the distributions of each centroid in the same class across different domains as follow:
	\begin{equation}
	\text{MMD}_c^2[\mathcal{D}_s,\mathcal{D}_t]\!=\!\!\sum_{l=1}^7\left|\left|\frac{1}{N_s^l}\!\sum_{i=1}^{N_s^l}\phi\!\left(\bm{x}_{i,l}^{s}\right)-\frac{1}{N_t^{\hat{l}}}\!\sum_{i=1}^{N_t^{\hat{l}}}\phi\!\left(\bm{x}_{i,\hat{l}}^{t}\right)\right|\right|^2,
	\label{eq5}
	\end{equation} 
	where $\bm{x}_{i,l}^{s}$ and $\bm{x}_{i,\hat{l}}^{t}$ are the $i$-th instance with class label $l$ in the source and target domains respectively. $N_s^l$ and $N_t^{\hat{l}}$ are numbers of source and target instances in class $l$.  Through explicitly minimizing the distance of the distributions across domains for each class separately, we can ensure the invariance of class-conditional distribution between different domains.
	
	Besides, the additional benefit of this class-wise adaption is two fold. First, one of the open issues for the pseudo label is the validity of the predicted result based on other different domains. And the wrong information in the pseudo-labeled target domain may deteriorate the adaption performance. In our method, by aligning the centroid of each class, the effects of these correctly and wrongly labeled samples are neutralized together so that the negative influences conveyed by the false pseudo-labeled samples can be suppressed by the correct labeled ones.
	Second, by calculating the MMD distances across domains for each class independently, our method can guarantee that samples with the same label can be mapped nearby in the learned feature space. Thus the class discriminative of learned features can be enhanced.
	
	Likewise, combining Eq. \ref{eq4} and Eq. \ref{eq5}, the unbiased approximation to $\text{MMD}_c$ can be formulated as:
	\begin{equation}
	\begin{split}
	\text{MMD}_{c,u}^2[\mathcal{D}_s,\mathcal{D}_t]=\sum_{l=1}^7&\left[\frac{1}{N_s^l(N_s^l-1)}\sum_{i\neq j}^{N_s^l}k\left(\bm{x}_{i,l}^s,\bm{x}_{j,l}^s\right)\right.\\
	+&\frac{1}{N_t^{\hat{l}}(N_t^{\hat{l}}-1)}\sum_{i\neq j}^{N_t^{\hat{l}}}k\left(\bm{x}_{i,\hat{l}}^t,\bm{x}_{j,\hat{l}}^t\right)\\
	-&\left.\frac{2}{N_s^l N_t^{\hat{l}}}\sum_{i,j=1}^{N_s^l,N_t^{\hat{l}}}k\left(\bm{x}_{i,l}^s,\bm{x}_{j,\hat{l}}^t\right)\right].
	\end{split}
	\end{equation}
	


	
	\subsection{Optimization of the Network}
	Without any label information in the target domain, directly adapting deep CNN to the target domain via fine-tuning  remains impossible and utilizing CNN learned on the source domain to classify the target domain is prone to over-fitting. So we embed the  MMD metrics proposed in Sec. \ref{method} to the CNN architecture,  in which way the deep network can leverage the labeled-source data and also explore the underlying information of the unlabeled-target data, then the output features can be generalized and discriminated enough on both domains. 
	As related research ~\cite{yosinski2014transferable} has suggested, the network becomes more task-aimed as  the layer goes deeper and hence it will become difficult to directly transfer the learned feature to the target domain. Therefore, we utilize the output of the last feature extraction layer 
	as the RKHS for the MMD layer so as to regularize the learned representation to be invariable  to domain shift.
	
	In addition, as the ground truth label $Y^t$ of the target domain is unavailable in this unsupervised circumstance, we then replace the ground truth with pseudo labels $\hat{Y^t}$ learned by the network in a fixed number of iterations. Different from previous methods that all pseudo labels of the target data have been directly used to calculate the weight ratio, we further introduce a parameter $\delta(l)$ to indicate the confidence of assigning $l$ as a tentative label to $\bm{x}^t$. A minimum entropy regularization  method \cite{grandvalet2005semi} has been adopted to learn  $\delta_i(l)$ for each $\bm{x}_i^t$: $\delta_i(l)=p(l|\bm{x}_i^t;\theta^k)$, where $\theta^k$ is the learned network parameters in the $k^{th}$ iteration. 
	In this way, we can take advantage of the unlabeled target data while regulating their contribution to provide robustness to the learning scheme. We then estimate the weigh ratio $\alpha$ as:
	\begin{equation}
	\label{eq10}
	\alpha_l=\frac{P(\hat{Y^t}=l)}{P(Y^s=l)}=\frac{\sum_{i=1}^{N_t}\delta_i(l)/N_t}{N_s^l/N_s},
	\end{equation}
	where $N_s^l$ are the number of samples with class $l$ in the source domain. By adding the class re-weighted MMD loss layer and the class-conditional MMD loss layer into the network, the objective function can be formulated as:
	\begin{equation}
	\label{eq11}
	\begin{split}
	L=\frac{1}{N_s}\sum_{i=1}^{N_s}L_s(\Theta(\bm{x}_i^s),y_i^s)&+\gamma \text{MMD}_{\alpha,u}^2[\mathcal{D}_s,\mathcal{D}_t]\\
	&+\lambda\text{MMD}_{c,u}^2[\mathcal{D}_s,\mathcal{D}_t],\\
	\end{split}
	\end{equation}
	where $L_s$ denotes the softmax loss of the source domain, $\Theta$ is the network parameters to be learned, $\gamma$ and $\lambda$ is the hyper-parameter to weight against these loss functions. In addition, we suppress the noisy information of pseudo labels by assigning a small weight to $\gamma$ and $\lambda$ in the early training phase. 
	The  gradients of the re-weighted MMD loss with respect to source feature $\bm{x}_i^s$ and target feature $\bm{x}_i^t$ can be computed as:
	\begin{equation}
	\label{eq12}
	\begin{split}
	\frac{\partial\text{MMD}_{\alpha,u}^2[\mathcal{D}_s,\mathcal{D}_t] }{\partial \bm{x}_i^s}&=\frac{\alpha_{y_i^s}\alpha_{y_j^s}}{N_s(N_s-1)}\sum_{i\neq j}^{N_s}\frac{\partial k(\bm{x}_i^s,\bm{x}_j^s)}{\partial \bm{x}_i^s}\\
	&-\frac{2 \alpha_{y_i^s}}{N_s N_t}\sum_{i,j=1}^{N_s,N_t}\frac{\partial k(\bm{x}_i^s,\bm{x}_j^t)}{\partial \bm{x}_i^s},
	\end{split}
	\end{equation}
	\begin{equation}
	\label{eq13}
	\begin{split}
	\frac{\partial\text{MMD}_{\alpha,u}^2[\mathcal{D}_s,\mathcal{D}_t] }{\partial \bm{x}_i^t}&=\frac{1}{N_t(N_t-1)}\sum_{i\neq j}^{N_t}\frac{\partial k(\bm{x}_i^t,\bm{x}_j^t)}{\partial \bm{x}_i^t}\\
	&-\frac{2 \alpha_{y_i^s}}{N_s N_t}\sum_{i,j=1}^{N_t,N_s}\frac{\partial k(\bm{x}_j^s,\bm{x}_i^t)}{\partial \bm{x}_i^t}.
	\end{split}
	\end{equation}
	And the  gradients of the class-conditional MMD loss with respect to source feature $\bm{x}_i^s$ and target feature $\bm{x}_i^t$ can be computed as:
	\begin{equation}
	\label{eq14}
	\begin{split}
	\frac{\partial\text{MMD}_{c,u}^2[\mathcal{D}_s,\mathcal{D}_t] }{\partial \bm{x}_i^s}=\sum_{l=1}^7&\left[\frac{1}{N_s^l(N_s^l-1)}\sum_{i\neq j}^{N_s^l}\frac{\partial k\left(\bm{x}_{i,l}^s,\bm{x}_{j,l}^s\right)}{\partial \bm{x}_i^s}\right.\\
	-&\left.\frac{2}{N_s^l N_t^{\hat{l}}}\sum_{i,j=1}^{N_s^l,N_t^{\hat{l}}}\frac{\partial k\left(\bm{x}_{i,l}^s,\bm{x}_{j,\hat{l}}^t\right)}{\partial \bm{x}_i^s}\right],
	\end{split}
	\end{equation}
	\begin{equation}
	\label{eq15}
	\begin{split}
	\frac{\partial\text{MMD}_{c,u}^2[\mathcal{D}_s,\mathcal{D}_t] }{\partial \bm{x}_i^t}=\sum_{l=1}^7&\left[\frac{1}{N_t^{\hat{l}}(N_t^{\hat{l}}-1)}\sum_{i\neq j}^{N_t^{\hat{l}}}\frac{\partial k\left(\bm{x}_{i,\hat{l}}^t,\bm{x}_{j,\hat{l}}^t\right)}{\partial \bm{x}_i^t}\right.\\
	-&\left.\frac{2}{N_s^l N_t^{\hat{l}}}\sum_{i,j=1}^{N_t^{\hat{l}},N_s^l}\frac{\partial k\left(\bm{x}_{j,l}^s,\bm{x}_{i,\hat{l}}^t\right)}{\partial \bm{x}_i^t}\right].
	\end{split}
	\end{equation}
	Given the Gaussian multi-kernel defined in Eq. (\ref{eq3}), we typically take $\partial k(\bm{x}_i^s,\bm{x}_j^t)/{\bm{x}_i^s}$ for example:
	\begin{equation}
	\frac{\partial k(\bm{x}_i^s,\bm{x}_j^t)}{\partial \bm{x}_i^s}=-\sum_{u=1}^d\frac{\beta_u}{\sigma_u^2}k_u(\bm{x}_i^s,\bm{x}_j^t)*(\bm{x}_i^s-\bm{x}_j^t).
	\end{equation}
	\begin{algorithm}[t]
		\small
		\caption{ Optimization algorithm of ECAN.}
		\label{al1}
		\begin{algorithmic}[1] 
			\REQUIRE ~~
			Source data \( \{ {(\bm{x}_i^s,y_i^s)}\} _{i = 1}^{n_s}\), Target data $\{\bm{x}_i^t\}_{i=1}^{n_t}$, \\\qquad\quad$n_s$ and $n_t$ is the size of mini-batch
			\ENSURE ~~ 
			Network layer parameters $\Theta$
			\STATE \textbf{Initialize:} The number of iteration $k\leftarrow0$, pseudo label's update interval $N_p$, Network learning rate $\mu$, hyper parameter $\gamma$ and $\lambda$, Network layer parameters \textbf{$\Theta$}.
			\WHILE{not converge}
			\STATE
			$k\leftarrow k+1$ 
			\IF{$k$ mod $N_p=0$}
			\STATE
			Update the pseudo label $\hat{y}_i^{t^{(k)}}$ and parameter $\delta_i(l)^{(k)}$ of target data $\bm{x}_i^{t^{(k)}}$.\\
			\ENDIF
			\STATE
			Compute the weight ratio $\alpha_l^k$ for the re-weighted MMD loss by Eq. \ref{eq10}.
			\STATE
			Compute the joint loss by Eq. \ref{eq11}: \\
			\begin{displaymath}
			L^k=L_s^k+\gamma L_{\text{MMD}}^{\alpha,u^{(k)}} + \lambda L_{\text{MMD}}^{c,u^{(k)}}
			\end{displaymath}
			\STATE
			Compute the back propagation error for each samples $\bm{x}_i$ by Eq. \ref{eq12}--\ref{eq15}:
			\begin{equation*}
			\begin{split}
			\frac{\partial L^k}{\partial \bm{x}_i^{s^{(k)}}}&=\frac{\partial L_s^k}{\partial \bm{x}_i^{s^{(k)}}}+\gamma\frac{\partial L_{\text{MMD}}^{\alpha,u^{(k)}}}{\partial \bm{x}_i^{s^{(k)}}}+\lambda\frac{\partial L_{\text{MMD}}^{c,u^{(k)}}}{\partial \bm{x}_i^{s^{(k)}}}\\
			\frac{\partial L^k}{\partial \bm{x}_i^{t^{(k)}}}&=\gamma\frac{\partial L_{\text{MMD}}^{\alpha,u^{(k)}}}{\partial \bm{x}_i^{t^{(k)}}}+\lambda\frac{\partial L_{\text{MMD}}^{c,u^{(k)}}}{\partial \bm{x}_i^{t^{(k)}}}
			\end{split}
			\end{equation*}
			\STATE
			Update the network layer parameters $\Theta$: 
			\begin{equation*}
			\begin{split}
			\Theta^{k+1}&=\Theta^k-\mu ^k\frac{\partial L^k}{\partial \Theta^k}\\
			&=\Theta^k-\mu ^k(\sum_{i=1}^{n_s}\frac{\partial L^k}{\partial \bm{x}_i^{s^{(k)}}}\frac{\partial \bm{x}_i^{s^{(k)}}}{\partial \Theta^k}+\sum_{i=1}^{n_t}\frac{\partial L^k}{\partial \bm{x}_i^{t^{(k)}}}\frac{\partial \bm{x}_i^{t^{(k)}}}{\partial \Theta^k})
			\end{split}
			\end{equation*}
			\ENDWHILE
		\end{algorithmic}
	\end{algorithm}
	
	By updating the network parameters with mini-batch SGD, the deep emotion-conditional adaption network (ECAN) can learn discriminative representations by utilizing information from the source domain, and in the meantime, matching both the marginal distribution and the conditional distribution between source and target domains so that the good separability among different expressions can also generalize well to the target data. What's more, by minimizing the differences in prior class probabilities, the ECAN can remedy the imbalanced problem in facial expression recognition. 
	Algorithm~\ref{al1} summarizes the learning process in the proposed  ECAN.
	
	
		\begin{figure}
		\small
		\centering
		\begin{minipage}{2.1cm}
			\centerline{\includegraphics[width=2.1cm]{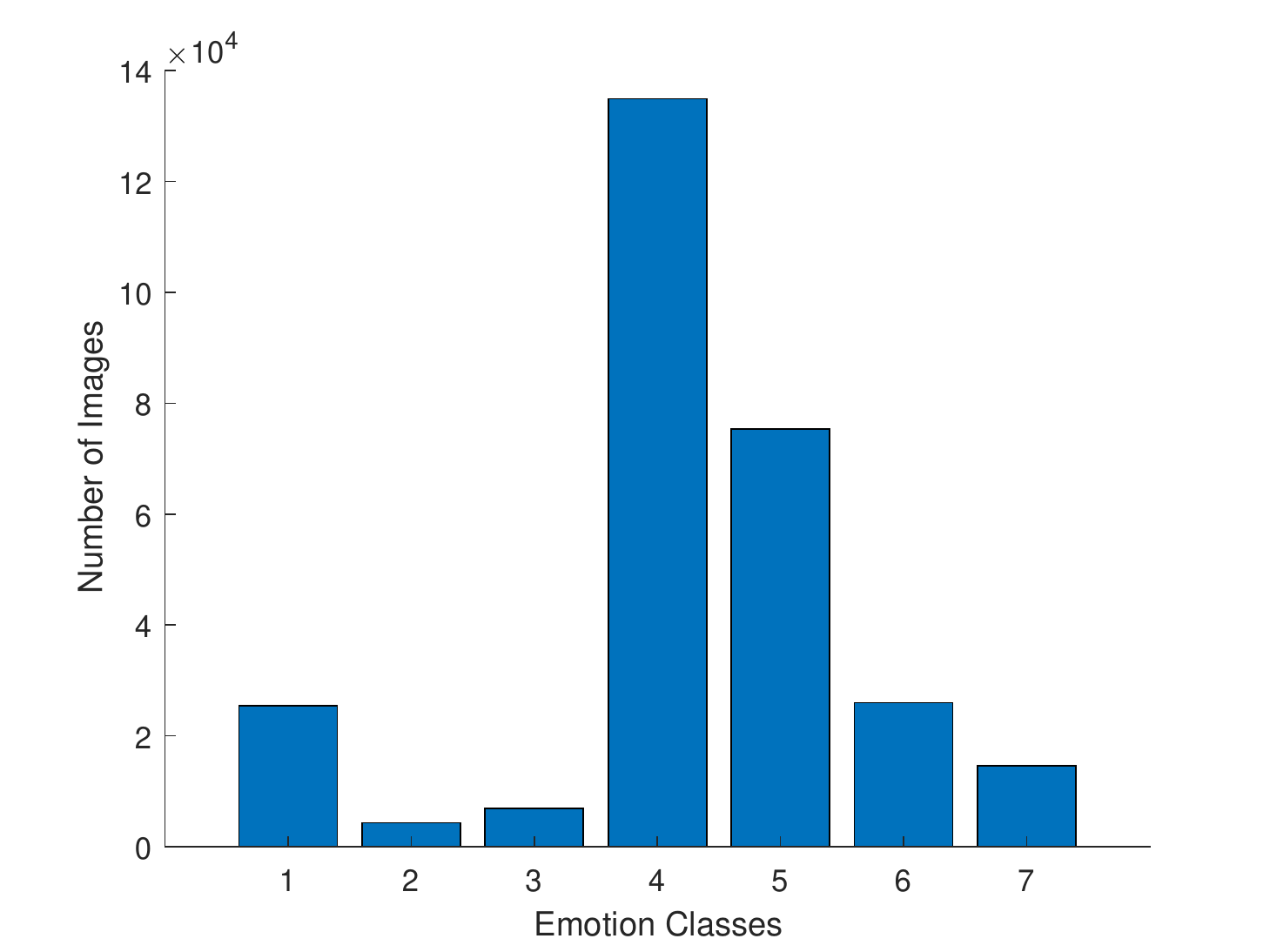}}
			\centerline{(a) AffectNet}  
		\end{minipage}
		\begin{minipage}{2.1cm}
			\centerline{\includegraphics[width=2.1cm]{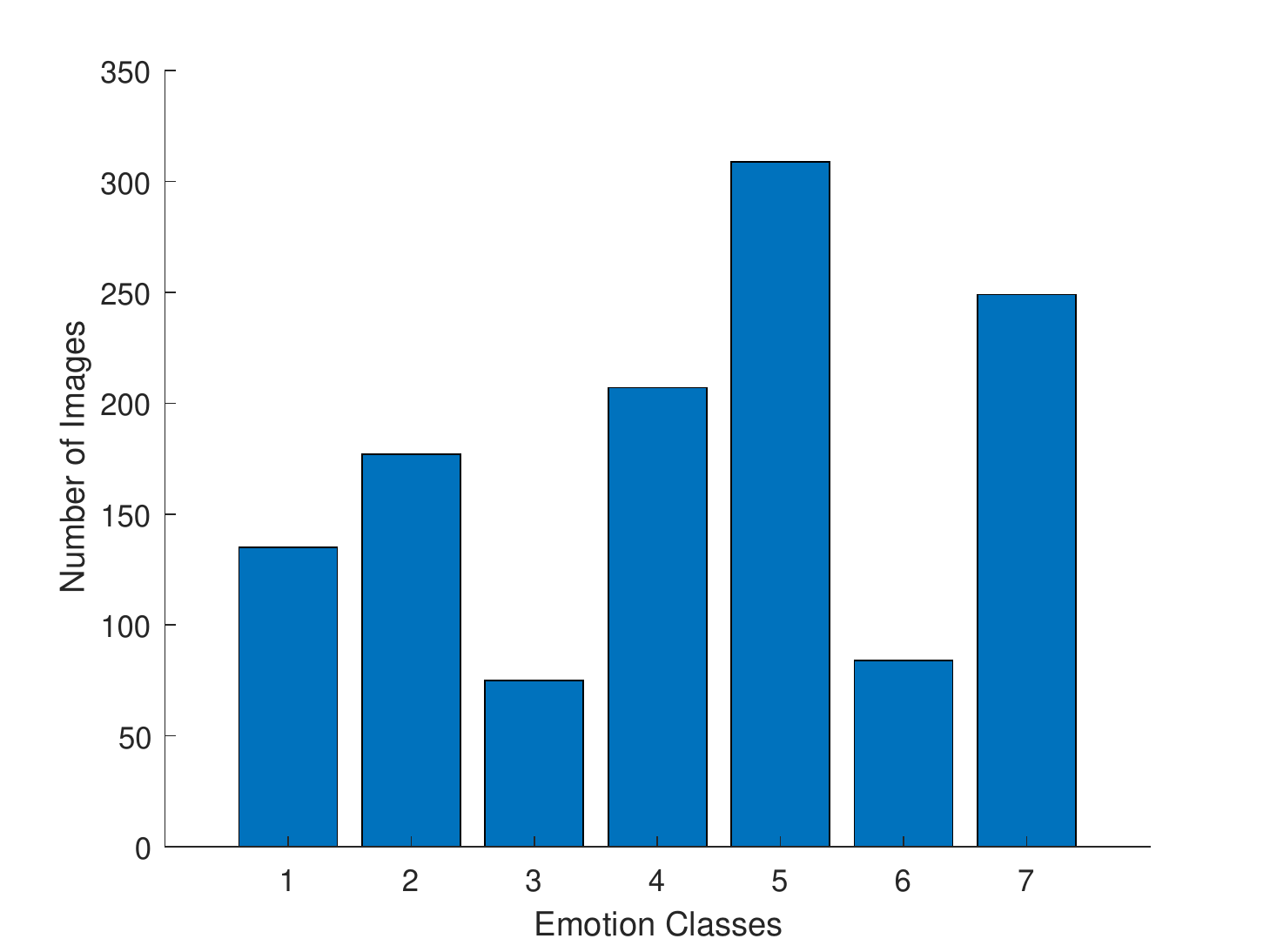}}
			\centerline{(b) CK+}  
		\end{minipage}
		\begin{minipage}{2.1cm}
			\centerline{\includegraphics[width=2.1cm]{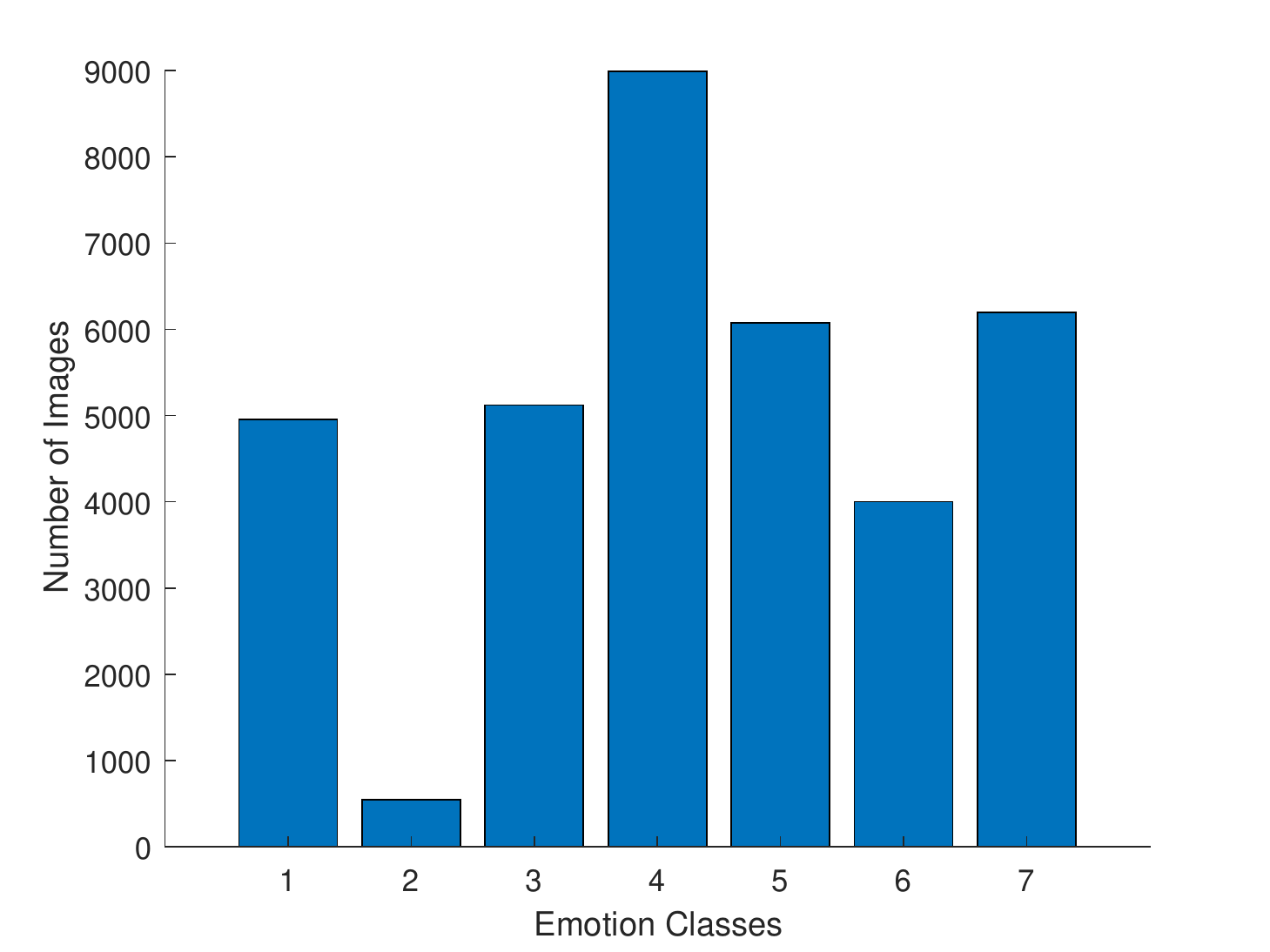}}
			\centerline{(c) FER2013}
		\end{minipage}	
		\begin{minipage}{2.1cm}
			\centerline{\includegraphics[width=2.1cm]{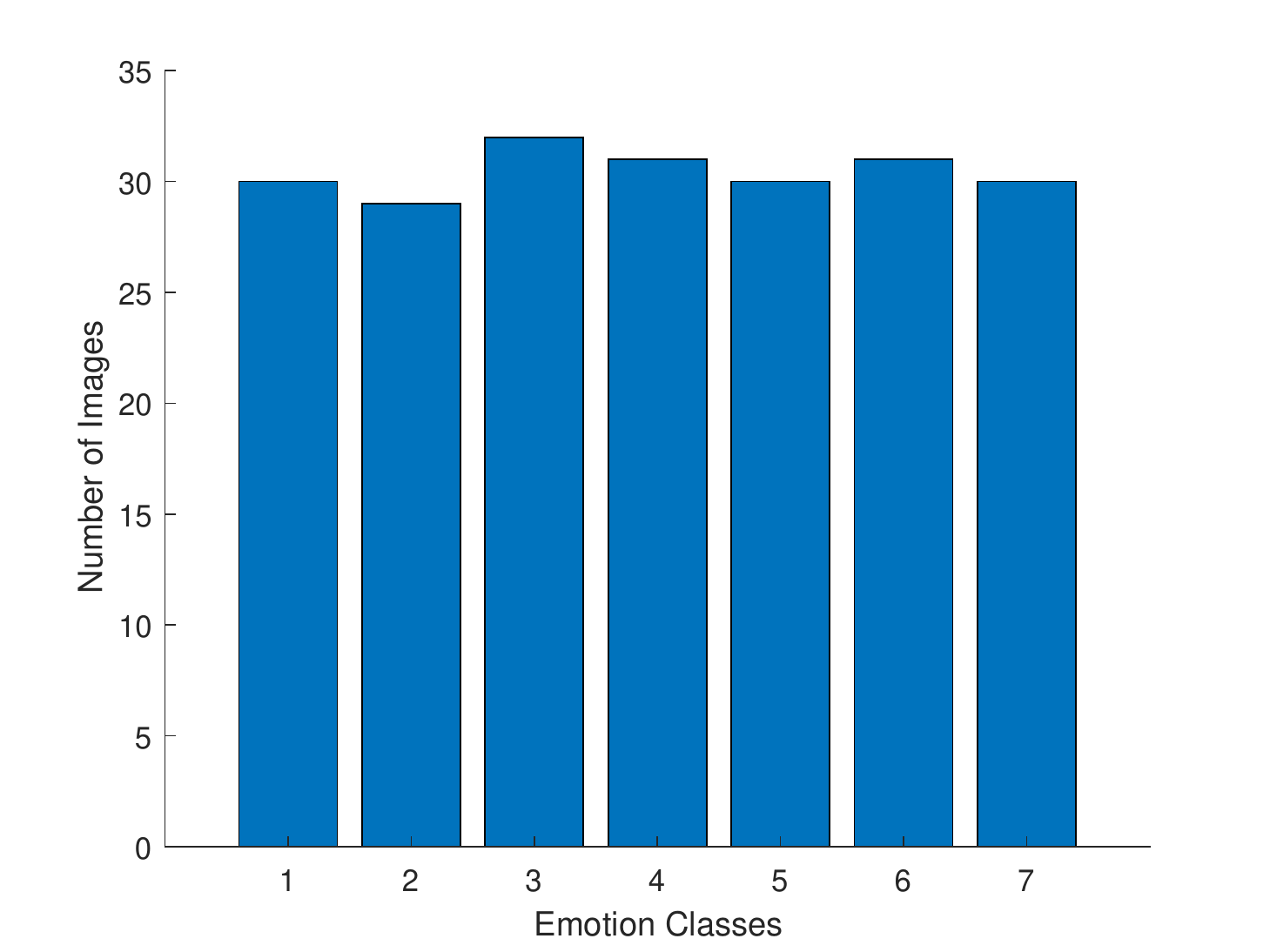}}
			\centerline{(d) JAFFE}
		\end{minipage}	
		
		\begin{minipage}{2.1cm}
			\centerline{\includegraphics[width=2.1cm]{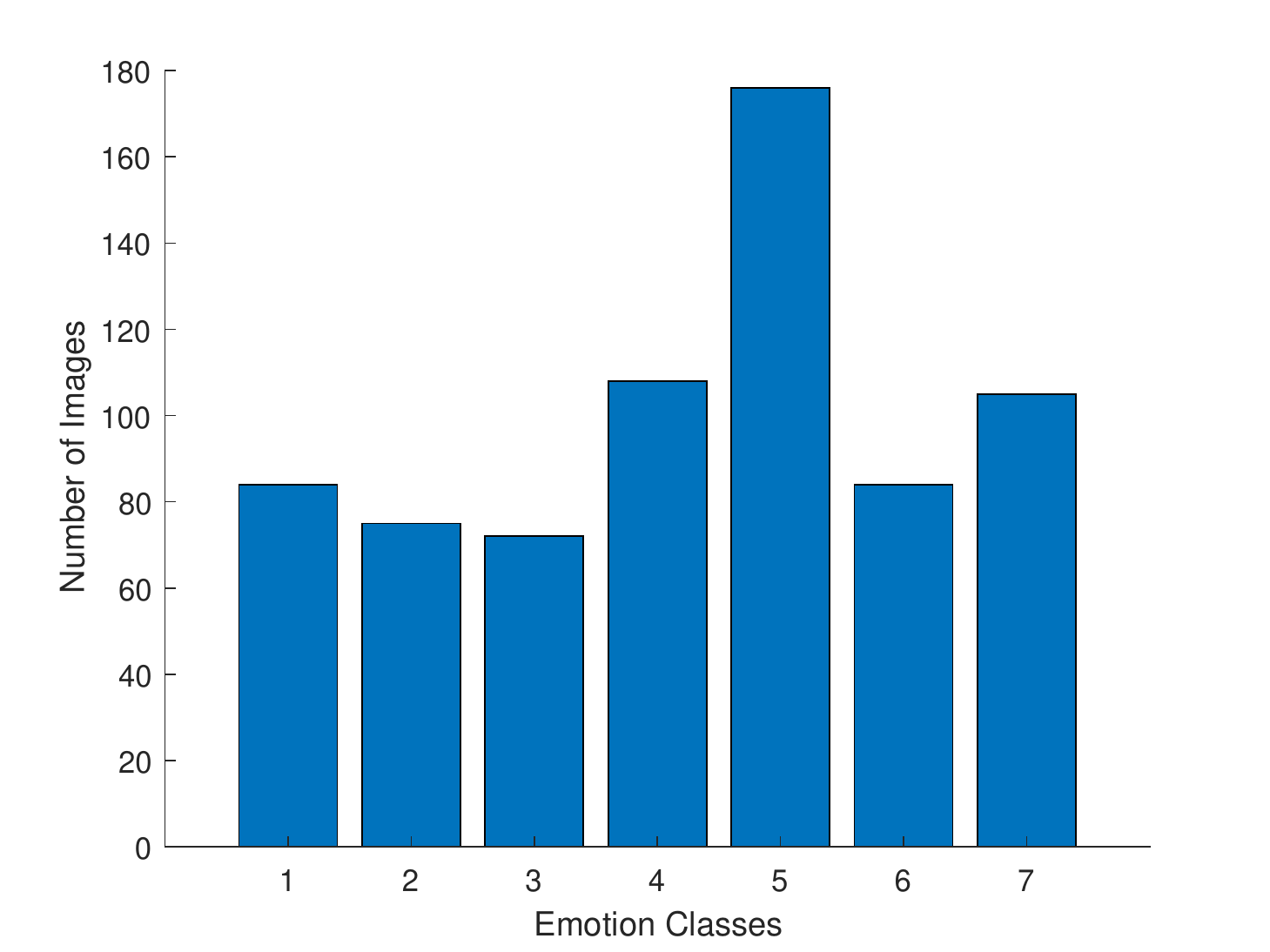}}
			\centerline{(e) MMI}
		\end{minipage}
		\begin{minipage}{2.1cm}
			\centerline{\includegraphics[width=2.1cm]{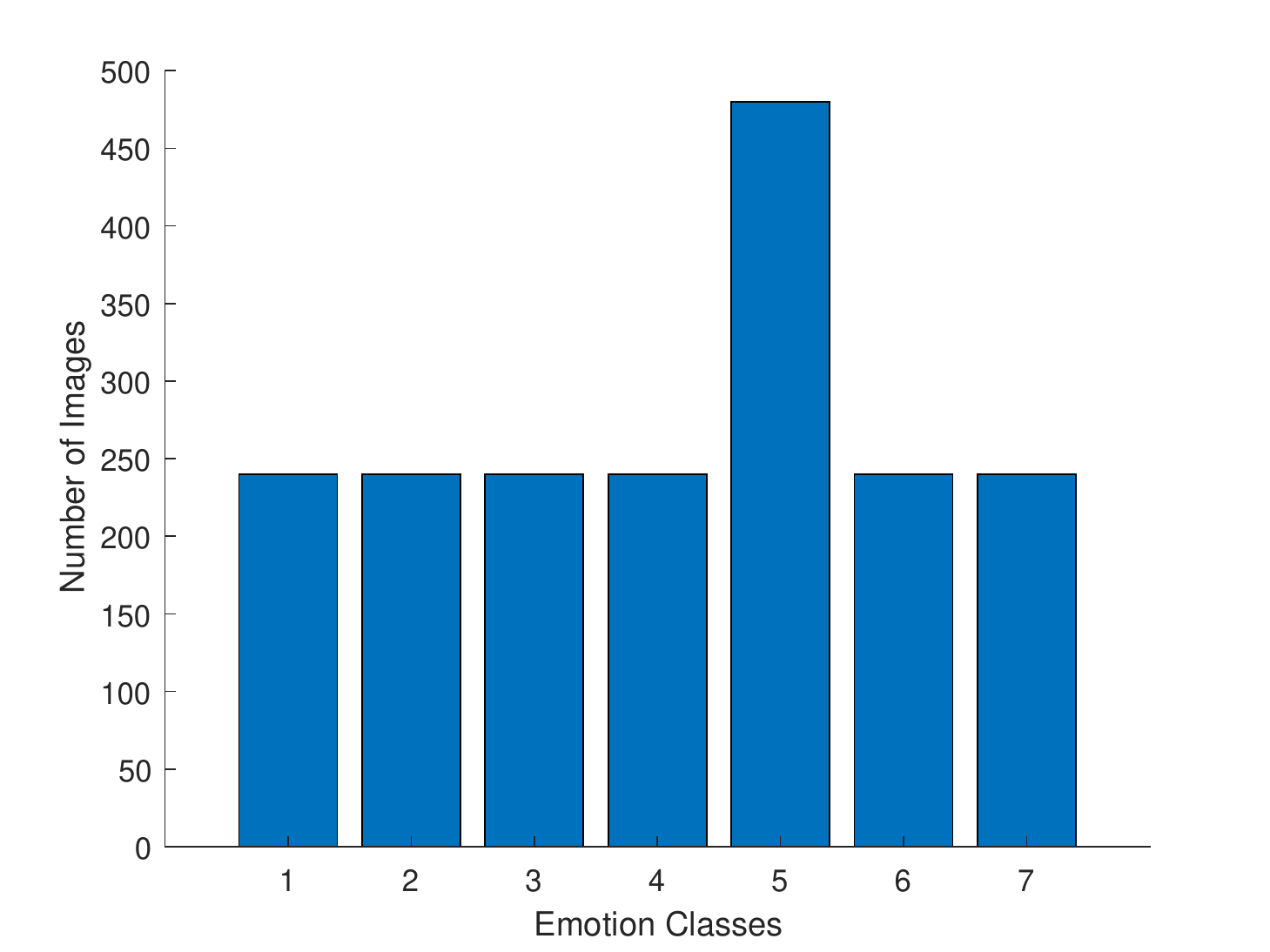}}
			\centerline{(f) Oulu-CASIA}
		\end{minipage}
		\begin{minipage}{2.1cm}
			\centerline{\includegraphics[width=2.1cm]{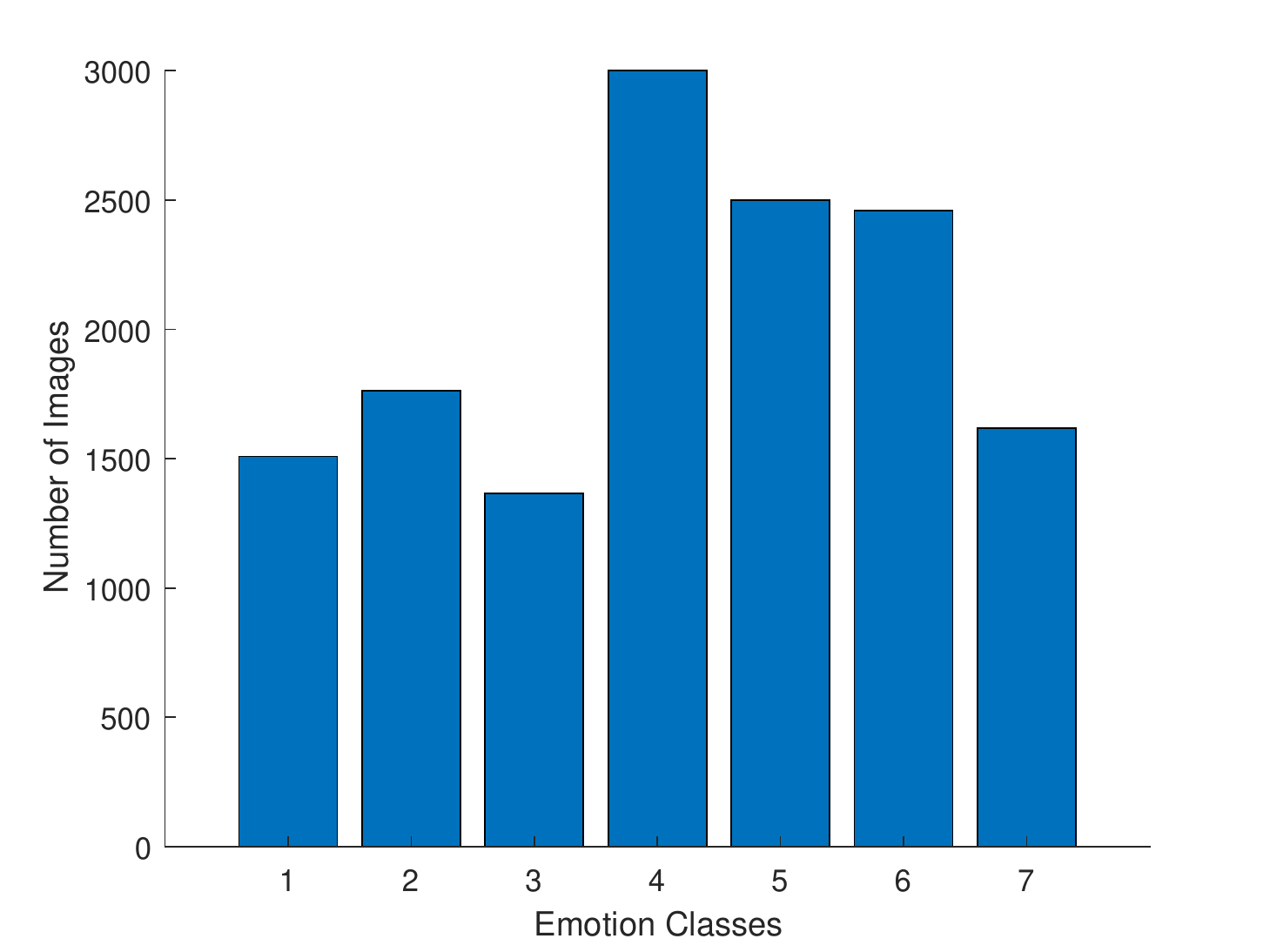}}
			\centerline{(g) RAF-DB 2.0}
		\end{minipage}
		\begin{minipage}{2.1cm}
			\centerline{\includegraphics[width=2.1cm]{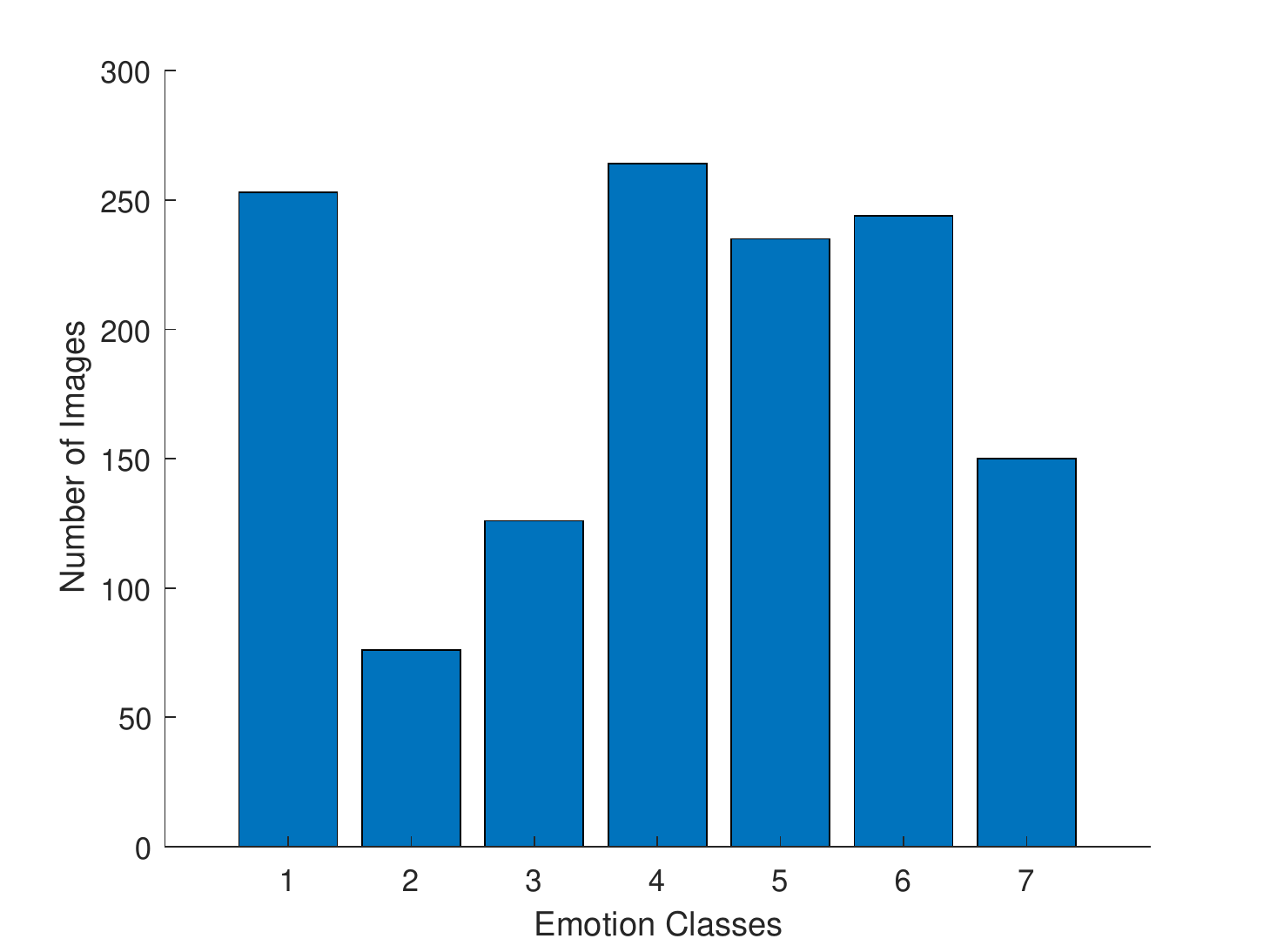}}
			\centerline{(h) SFEW 2.0}
		\end{minipage}
		\caption{Distribution of seven expressions for seven representative databases. 1, 2, 3, 4, 5, 6, 7 stand for Anger, Disgust, Fear, Happiness, Neutral, Sadness, Surprise respectively. We can see that class imbalance problem is very common in facial expression datasets, especially for real-world scenarios. Note that for the lab-controlled dataset JAFFE and Oulu-CASIA, the number of each category is intentionally set to be the same by experimenters.}
		\label{dis}
	\end{figure}
	
	\section{experiments and results}
	\label{sec:experiment}
	In this section, we will conduct extensive experiments  to evaluate the proposed unsupervised cross-dataset facial expression recognition method. Four lab-controlled datasets (CK+, JAFFE, MMI and Oulu-CASIA) and four real-world datasets (AffectNet, FER2013, RAF-DB 2.0 and SFEW 2.0) are used to examine the adaption power of our method.
	\subsection{Databases and Setups}
	AffectNet, CK+, FER2013, JAFFE, MMI, Oulu-CASIA, RAF-DB and SFEW 2.0 are widely-used facial expression datasets. Fig. \ref{fig:bias} shows the example samples of these datasets. And the class distributions of these datasets used in the experiments are shown in Fig. \ref{dis}. 
	
	(1) AffectNet Database~\cite{Mollahosseini2017AffectNet}: The AffectNet database contains around one million facial images downloaded from the Internet by querying different search engines using 1,250 emotion related tags in different languages. During experiment, around 280,000 images from the training set and 3,500 images from the validation set with neutral and six basic expression labels are chosen.
	
	(2) CK+ Database~\cite{lucey2010extended}: The lab-controlled database CK+ contains 593 video sequences from 123 subjects. Only 309 sequences have been labeled with six basic expression labels (excluding Neutral). We then extract the final three frames of each sequence with peak formation, and in the meanwhile select the first frame (neutral face) from 309 sequences, resulting in 1,236 images. 
	
	(3) FER2013 Database~\cite{goodfellow2013challenges}: The large-scale and unconstrained database FER2013 was created and labeled automatically by the Google image search API. All images in FER2013 have been registered and resized to $48\times48$ pixels. We  use  35,887 images with expression labels during experiments.
	
	(4) JAFFE Database~\cite{lyons1998japanese}: The JAFFE is a laboratory-controlled database which contains only 213 samples from 10 Japanese females with posed expressions. Each person has 3˜4 images with each of six basic facial expressions and one image with a neutral expression. All images have been used in our experiments.
	
	(5) MMI Database~\cite{valstar2010induced}: The MMI is a lab-controlled database which includes 2,900 video sequences from 75 subjects with non-uniformly posed expressions and various accessories. We select the first frame (neutral face) and the three peak frames in each sequence with expression labels, resulting in 704 images.
	
	(6) Oulu-CASIA Database~\cite{zhao2011facial}: The lab-controlled Oulu-CASIA database includes 2,880 image sequences collected from 80 subjects labeled with six basic emotion labels. Each of the videos is captured with one of two imaging systems, i.e., near-infrared (NIR) or visible light (VIS), under three different illumination conditions. Similar to CK+, we select the last three peak frames and the first frame (neutral face) from the 480 videos with the VIS System under normal indoor illumination, resulting in 1,920 images.
	
	(7) RAF-DB 2.0 Database \footnote{http://www.whdeng.cn/RAF/model1.html}: The RAF-DB 2.0 is an extension to the current RAF-DB database \cite{li2017reliable,li2018reliable}. RAF-DB is a large-scale dataset which contains about 30,000 great diverse facial images from thousands of individuals downloaded from the Internet. To address the concerns on the
	imbalanced distribution of expression categories, we augment the dataset with more samples in rare expression classes.
	Following the previous collection criteria, 642 images with
	anger label, 886 images with disgust label and 1,010 images
	with fear label have been supplemented into the original
	datasets to form a more balanced dataset, RAF-DB 2.0. We then choose a subset of the database with basic emotions for the experiments, in total 14,216 images.
	
	
	(8) SFEW 2.0 Database~\cite{dhall2011static}: The in-the-wild database SFEW contains dynamic images selected from different films with spontaneous expressions, various head pose, age range, occlusions and illuminations. The challenging database is divided into three sets for training, validation and testing. During experiments, we use 921 images in the training set and 427 images in the validation part provided with labels, in total 1,358 images. 
	
\begin{figure*}[htb]
		\centering
		\subfigure[AffectNet look-alikes from RAF-DB 2.0]{
			\includegraphics[height=3.5cm]{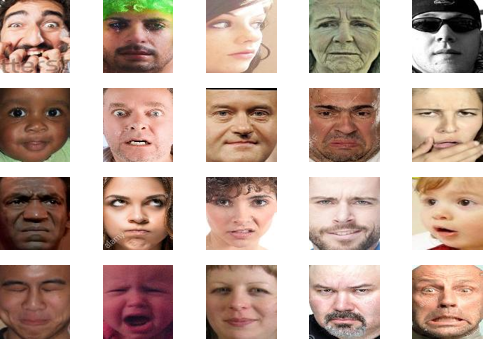}}
		\hspace{.3in}
		\subfigure[CK+ look-alikes from RAF-DB 2.0]{
			\includegraphics[height=3.5cm]{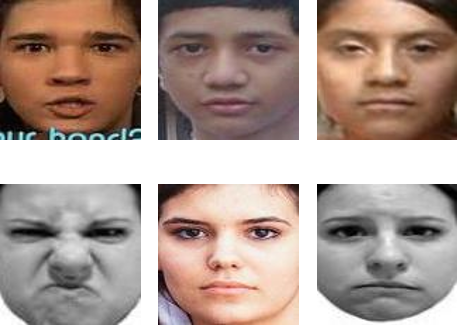}}
		\hspace{.3in}
		\subfigure[FER2013 look-alikes from RAF-DB 2.0]{
			\includegraphics[height=3.5cm]{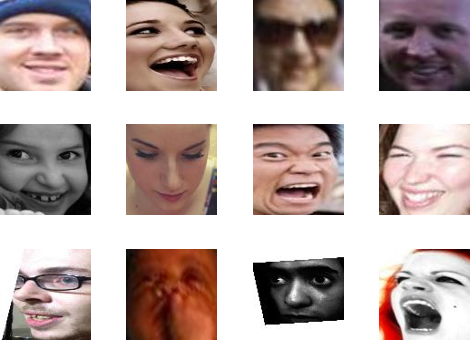}}
		\subfigure[MMI look-alikes from RAF-DB 2.0]{
			\includegraphics[height=3.5cm]{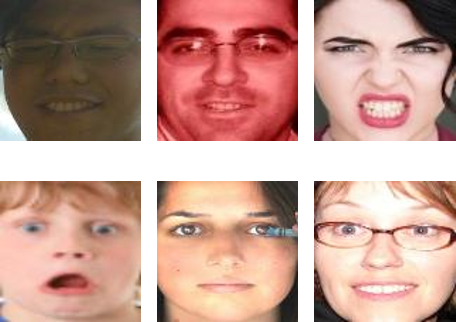}}
		\hspace{.3in}
		\subfigure[Oulu-CASIA look-alikes from RAF-DB 2.0]{
			\includegraphics[height=3.5cm]{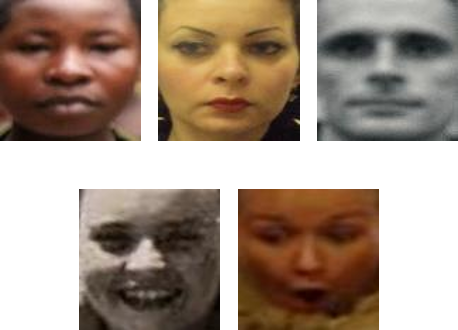}}
		\hspace{.3in}
		\subfigure[SFEW 2.0 look-alikes from RAF-DB 2.0]{
			\includegraphics[height=3.5cm]{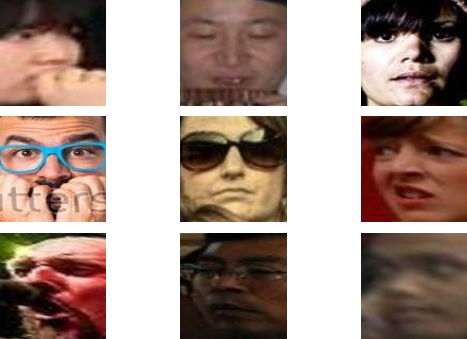}}
		\caption{Other datasets Look-alike from RAF-DB 2.0. Samples from RAF-DB 2.0 that are closest to the decision boundaries of SVM trained in the dataset recognition experiment are displayed.}
		\label{fig:alike}
	\end{figure*}

	As RAF-DB 2.0 dataset achieves the best cross-dataset performance on the seven expressions classification task (see the mean others values in Table \ref{tab:cross}), we choose RAF-DB 2.0 as the source domain in our single-source domain adaption experiments. In Figure \ref{fig:alike}, we further visualize the other datasets look-alike images from RAF-DB 2.0 by picking out samples that are closed to the decision boundary, to see how RAF-DB 2.0 can resemble other different datasets.

	\subsection{Implementation Details}
	In our experiments, we finetune our model based on the VGG-Face architecture \cite{simonyan2014very} that pre-trained on a large-scale face dataset with 2.6M images. Two MMD-based regularization layers are appended to the second to last fully-connected layer and the dimension of the last label prediction layer is amended to 7 for facial expression recognition. 
	The already aligned gray-scale images of each dataset are firstly resized to $256\times256$ and then random $224\times224$ pixel patches cropped from these images are fed into the network. We also augment the data by flipping it horizontally with 50\% probability. 
	
	For hype-parameters settings, no labeled target samples are referred in our unsupervised domain adaptation experiments.  And we  tune the trade-off parameters $\gamma$ and $\lambda$ from the sets $\{ 0, 0.01, 0.03, 0.05, 0.1, 0.3, 0.5, 1\}$ and $\{0,0.001,0.01,0.1\}$ by grid searching, respectively. To suppress the noisy signal from the pseudo label of target data, we further assign a relatively small weight $w$ to these two parameters at the early stage of the training process. Specifically, $w=\frac{2}{1+exp(-10\cdot p)}-1$, where $p$ is the training progress linearly changing from 0 to 1.
	During training, stochastic gradient descent with 0.9 momentum is used. And the based learning rate is set to 0.001. As the last classifier layer is trained from scratch, we set its learning rate to be 10 times that of the lower layers. All experiments are implemented by Caffe Toolbox \cite{jia2014caffe}, and run on a PC with a NVIDIA GTX 1080 GPU.

	\subsection{Experimental Results}
	In this subsection, we comprehensively evaluate our deep domain adaption method on cross-dataset facial expression recognition tasks in terms of the classification accuracy to demonstrate the effectiveness of our proposed algorithm.
	The cross-domain results on four lab-controlled databases and comparisons with other state-of-the-arts are shown in Table. \ref{CK+} -- Table. \ref{oulu}. The comparison results of the other three real-world databases are shown in Table. \ref{sfew}. Note that ``CNN'' refers to the VGG-Face network directly fine-tuned on the source dataset and ``CNN+MMD'' refers to the original MMD that only match the marginal distribution across domain. And we list their best results after parameter tuning.
	
	On the whole, despite that different methods may employ different datasets as the source domain, we can still observe that our network outperforms the comparison methods on most datasets, and can achieve competitive performances on difficult transfer tasks, especially when source and target  domains are much more different. We then analyze the results for each dataset in detail.
	\begin{table}[!t]
		\centering
		\renewcommand\arraystretch{1.05}
		\setlength{\tabcolsep}{5pt}
		\caption{Comparison of our methods with other results in the literature on CK+ dataset.}
		\label{CK+}
		\begin{threeparttable}
			\begin{tabular*}{8.5cm}{@{}cccc@{}}
				\toprule
				\multicolumn{2}{c}{Methods} & Source Dataset & Accuracy \\ \midrule
				\multirow{5}{*}{\begin{tabular}[c]{@{}c@{}}Shallow\\ Models\end{tabular}} & Da et al.~\cite{da2015effects} & BOSPHORUS & 57.60\% \\
				& Zhang et al.~\cite{zhang2015facial} & MMI & 61.20\% \\
				& Miao et al.~\cite{miao2012cross} & MMI + JAFFE & 65.0\% \\
				& Lee et al.~\cite{lee2014intra} & MMI & 65.47\%\\
				& Mayer et al.~\cite{mayer2014cross} & MMI & 66.20\% \\\hline
				\multirow{5}{*}{\begin{tabular}[c]{@{}c@{}}Deep \\ Models\end{tabular}} & Mollahosseini~\cite{mollahosseini2016going} & 6 Datasets\tnote{$\dagger$} & 64.2\% \\
				& Zavarez et al.~\cite{zavarez2017cross} & 6 Datasets\tnote{$\ddagger$} & 88.58\% \\
				& Hasani et al.~\cite{hasani2017spatio} & MMI + JAFFE & 73.91\% \\
				& Wen et al.~\cite{wen2017ensemble} & FER2013 & 76.05\% \\
				&Wang et al.~\cite{wang2018unsupervised}&FER2013&76.58\%\\\hline
				\multirow{4}{*}{\begin{tabular}[c]{@{}c@{}}Our\\ Methods\end{tabular}}  & CNN & RAF-DB 2.0 &  78.00\%\\
				& CNN + MMD & RAF-DB 2.0 & 82.44\% \\
				& ECAN & RAF-DB 2.0 &86.49\%  \\ \cline{2-4}
				& ECAN & 4 Datasets\tnote{$\star$} &\textbf{89.69\% } \\\bottomrule
			\end{tabular*}
			\begin{tablenotes}
				\footnotesize
				\item[$\dagger$] MultiPIE, MMI, DISFA, FERA, SFEW, and FER2013.
				\item[$\ddagger$] JAFFE, MMI, RaFD, KDEF, BU3DFE and ARFace.
				\item[$\star$] RAF-DB 2.0, JAFFE, MMI, Oulu-CASIA
			\end{tablenotes}
		\end{threeparttable}
	\end{table}
	
	\begin{table}[!t]
		\centering
		\renewcommand\arraystretch{1.05}
		\setlength{\tabcolsep}{3pt}
		\caption{Comparison of our methods with other results in the literature on JAFFE dataset.}
		\label{jaf}
		\begin{threeparttable}
			\begin{tabular*}{8.8cm}{@{\extracolsep{\fill}}cccc@{}}
				\toprule
				\multicolumn{2}{c}{Methods} & Source Dataset & Accuracy \\ \midrule
				\multirow{5}{*}{\begin{tabular}[c]{@{}c@{}}Shallow\\ Models\end{tabular}} & Shan et al.~\cite{shan2009facial} & CK & 41.30\% \\
				& El et al.~\cite{el2014fully} & Bu-3DFE & 41.96\% \\
				& Da et al.~\cite{da2015effects} & CK+ & 42.30\% \\
				& Zhou et al.~\cite{zhou2013feature} & CK & 45.71\% \\
				&Gu et al.~\cite{gu2012facial} & CK+ & 55.87\% \\\hline
				\multirow{3}{*}{\begin{tabular}[c]{@{}c@{}}Deep \\ Models\end{tabular}} & Wen et al.~\cite{wen2017ensemble} & FER2013 & 50.70\% \\
				& Ali et al.~\cite{ali2016boosted} & RaFD & 48.67\% \\
				& Zavarez et al.~\cite{zavarez2017cross} & 6 datasets\tnote{$\star$} & 44.32\% \\\hline
				\multirow{3}{*}{\begin{tabular}[c]{@{}c@{}}Our\\ Methods\end{tabular}}  & CNN & RAF-DB 2.0 &  54.26\%\\
				& CNN + MMD & RAF-DB 2.0 &  58.64\%\\
				& ECAN & RAF-DB 2.0 & \textbf{61.94\%} \\ \bottomrule
			\end{tabular*}
			\begin{tablenotes}
				\footnotesize
				\item[$\star$] CK+, MMI, RaFD, KDEF, BU3DFE and ARFace.
			\end{tablenotes}
		\end{threeparttable}
	\end{table}
	
	\begin{table}[!t]
		\centering
		\renewcommand\arraystretch{1.05}
		\setlength{\tabcolsep}{3.5pt}
		\caption{Comparison of our methods with other results in the literature on MMI dataset.}
		\label{mmi}
		\begin{threeparttable}
			\begin{tabular*}{8.8cm}{@{\extracolsep{\fill}}cccc@{}}
				\toprule
				\multicolumn{2}{c}{Methods} & Source Dataset & Accuracy \\ \midrule
				\multirow{4}{*}{\begin{tabular}[c]{@{}c@{}}Shallow\\ Models\end{tabular}} & Shan et al.~\cite{shan2009facial} & CK & 51.10\% \\
				& Cruz et al.~\cite{cruz2014one} & CK+ & 57.6\% \\
				& Mayer et al.~\cite{mayer2014cross} & CK & 60.30\% \\
				& Zhang et al.~\cite{zhang2015facial} & CK+ & 66.90\% \\\hline
				\multirow{5}{*}{\begin{tabular}[c]{@{}c@{}}Deep \\ Models\end{tabular}} & Zavarez et al.~\cite{zavarez2017cross} & 6 datasets\tnote{$\dagger$} & 67.03\% \\
				& Mollahosseini~\cite{mollahosseini2016going} & 6 datasets\tnote{$\ddagger$} & 55.6\% \\
				& Hasani et al.~\cite{hasani2017facial} & CK+ & 54.76\% \\
				& Wang et al.~\cite{wang2018unsupervised}&FER2013&61.86\%\\
				& Hasani et al.~\cite{hasani2017spatio} & CK+ & 68.51\% \\\hline
				\multirow{3}{*}{\begin{tabular}[c]{@{}c@{}}Our\\ Methods\end{tabular}}  & CNN & RAF-DB 2.0 &64.13\%  \\
				& CNN + MMD & RAF-DB 2.0 &  65.80\%\\
				& ECAN & RAF-DB 2.0 &  \textbf{69.89\%}\\ \bottomrule
			\end{tabular*}
			\begin{tablenotes}
				\footnotesize
				\item[$\dagger$] CK+, JAFFE, RaFD, KDEF, BU3DFE and ARFace.
				\item[$\ddagger$] MultiPIE, CK+, DISFA, FERA, SFEW, and FER2013.
			\end{tablenotes}
		\end{threeparttable}
	\end{table}
	
	\begin{table}[t]
		\renewcommand\arraystretch{1.1}
		\setlength{\tabcolsep}{4.5pt}
		\centering
		\caption{Cross-dataset results on Oulu-CASIA dataset.}
		\label{oulu}
		\begin{tabular}{|c|c|c|c|}
			\hline
			Target&Methods&Source &Accuracy\\\hline
			\multirow{3}{*}{Oulu-CASIA}&CNN&\multirow{3}{*}{RAF-DB 2.0}&59.39\%\\
			&CNN + MMD&&60.14\%\\
			&ECAN&&\textbf{63.97\%}\\\hline
		\end{tabular}
	\end{table}
	
	For CK+ dataset, Our ECAN achieves much better performance than most other methods except \cite{zavarez2017cross} that conducted cross-dataset facial expression recognition combining six lab-controlled datasets as the source domain.
	It is worth noting that CK+ and these lab-controlled datasets are very similar in many respects, such as the controlled collection environment, subject characters, illumination condition and head postures. Hence methods in the literature that used these databases for cross-domain expression recognition can achieve relatively good performance on CK+. However, our method only adopt an unconstrained dataset which is collected from the Internet and much more diverse than the lab-controlled ones for the transfer tasks.  
	So we further evaluate our method using RAF-DB 2.0 with other three lab-controlled datasets as our multi-source domain, and achieve the best cross-dataset performance 89.69\%.
	On the other hand, when compared to the CK+ in-dataset results in Table \ref{cross} (85.40\%), ECAN also gains a slight performance improvement, which indicates that our method can mitigate the discrepancy across different datasets. 
	
	For JAFFE dataset, which is a highly biased dataset in respect of gender and ethnicity, i.e., it only contains ten Japanese females, the fine-tuning technique used in \cite{zavarez2017cross} which reported high accuracy on CK+ dataset is no longer effective in this context. In contrast, by matching the marginal and conditional distribution and also the class distribution across domains, our method yields the best performance and is superior to the highest accuracy of the literature \cite{gu2012facial} by 6.07\%.
	
	For MMI dataset, our ECAN achieves 69.89\% cross-dataset accuracy, which outperforms all the other compared methods and also the in-dataset performance shown in Table \ref{cross}. Comparing with the baselines, we can find that the original CNN structure is inferior than some previous methods that use more similar source dataset (such as CK+) with target MMI dataset and the original MMD (CNN+MMD) only gains a negligible improvement. However, with the help of the re-weighted MMD and the class-conditional MMD, the ECAN can ameliorate the effect of the class distribution bias and learn features with more discriminative ability, thus achieves superior results. 
	
	For Oulu-CASIA dataset, we can observe similar results: With the supervision of the two MMD regularizations, which match the class distributions and close both the marginal and conditional distribution distance across domains, our method can therefore achieve better performances when compared with the baselines and the in-dataset recognition performance. 
	
		\begin{table}[!t]
		\centering
		\renewcommand\arraystretch{1.1}
		\setlength{\tabcolsep}{4pt}
		\caption{Comparison of our methods with other results in the literature on AffectNet, FER2013 and SFEW 2.0 dataset.}
		\label{sfew}
		\begin{threeparttable}
			\begin{tabular}{|l|ccc|}
				\hline
				Target        & Method   & Source  & Accuracy \\ \hline
				\multirow{3}{*}{AffectNet}&CNN&RAF-DB 2.0&49.29\%\\
				&CNN + MMD&RAF-DB 2.0&48.76\%\\
				&ECAN&RAF-DB 2.0&\textbf{51.84\%}\\\hline\hline
				\multirow{7}{*}{SFEW} & El et al.~\cite{el2014fully} & Bu-3DFE & 20.57\% \\
				& Liu et al.~\cite{liu2015inspired} & CK+ & 29.43\% \\
				& Mollahosseini~\cite{mollahosseini2016going} & 6 datasets\tnote{$\dagger$} & 39.8\% \\\cline{2-4} 
				& CNN & RAF-DB 2.0 & 52.67\% \\
				& CNN + MMD & RAF-DB 2.0 &  52.81\%\\
				&ECAN & RAF-DB 2.0 & \textbf{54.34\%} \\ \hline\hline
				\multirow{4}{*}{FER2013}&Mollahosseini~\cite{mollahosseini2016going} & 6 datasets\tnote{$\ddagger$} & 34.0\% \\\cline{2-4} 
				& CNN & RAF-DB 2.0 & 55.38\% \\
				& CNN + MMD & RAF-DB 2.0 & 56.54\% \\
				&ECAN & RAF-DB 2.0 &\textbf{58.21\%}  \\ \hline
			\end{tabular}
			\begin{tablenotes}
				\footnotesize
				\item[$\dagger$] MultiPIE, CK+, DISFA, FERA, MMI, and FER2013.
				\item[$\ddagger$] MultiPIE, CK+, DISFA, FERA, MMI, and SFEW.
			\end{tablenotes}
		\end{threeparttable}
	\end{table}

	For AffectNet dataset, we find that the performance of the original MMD is slightly degraded due to the very different class distribution. However, our ECAN can mitigate this discrepancy and thus help boost the performance. And For the other two in-the-wild datasets FER2013 and SFEW 2.0, our method significantly outperforms the comparison methods. Firstly, because most previous methods have chosen lab-controlled databases that deviate from these two target sets as the source domain, our method which uses the similar in-the-wild RAF-DB 2.0 can perform better. Secondly, from the other perspective, as the domain discrepancy may not be distinct in our case and the original MMD provides limited improvements over CNN, our ECAN can further help enhance the performance by a certain margin, which demonstrates that the ECAN is able to transfer deep models across various domains. In addition, results on the large-scale test set FER2013 reveals that our method can also boost the recognition rate when the target domain size is much larger than the source domain size.
	
	Looking deeper into the experimental results, we can make the following observations: 
	(1) In consistent with the conclusion in Sec. \ref{sec:look}, well-trained deep learning based methods generally outperform most traditional shallow models for cross-database facial expression recognition, which is mainly due to their generalization characteristic and the capacity for learning discriminative representations. 
	(2) Among the deep learning methods, the performances of fine-tuning  technique suffer a large decline when domain discrepancy becomes obvious, which indicates that the challenge of domain discrepancy cannot be simply settled by fine-tuning techniques and mitigating the domain shift between different source and target datasets is of great significance for cross-dataset facial expression recognition. 
	(3) In some cases that domain discrepancy is relatively small or the class distribution bias is obvious, the original MMD can yield only a small gain or even worse performance. Nevertheless, our ECAN that considers both marginal and conditional distributions and also the class distribution bias can help improve the cross-dataset recognition accuracy by learning class discriminative feature representations and alleviating the class weight bias, which reveals its effectiveness in emotion transfer learning task varying different kinds of domains.
	
	\subsection{Empirical  Analysis}
	In this subsection, we further analyze the effectiveness of ECAN from several different aspects.
	\subsubsection{Parameter sensitivity}
\begin{figure}[t]
		\centering
		\subfigure[\bm{$\gamma$} on CK+ (left) / FER2013 (right)]{\label{par11}
			\includegraphics[width=2.75cm]{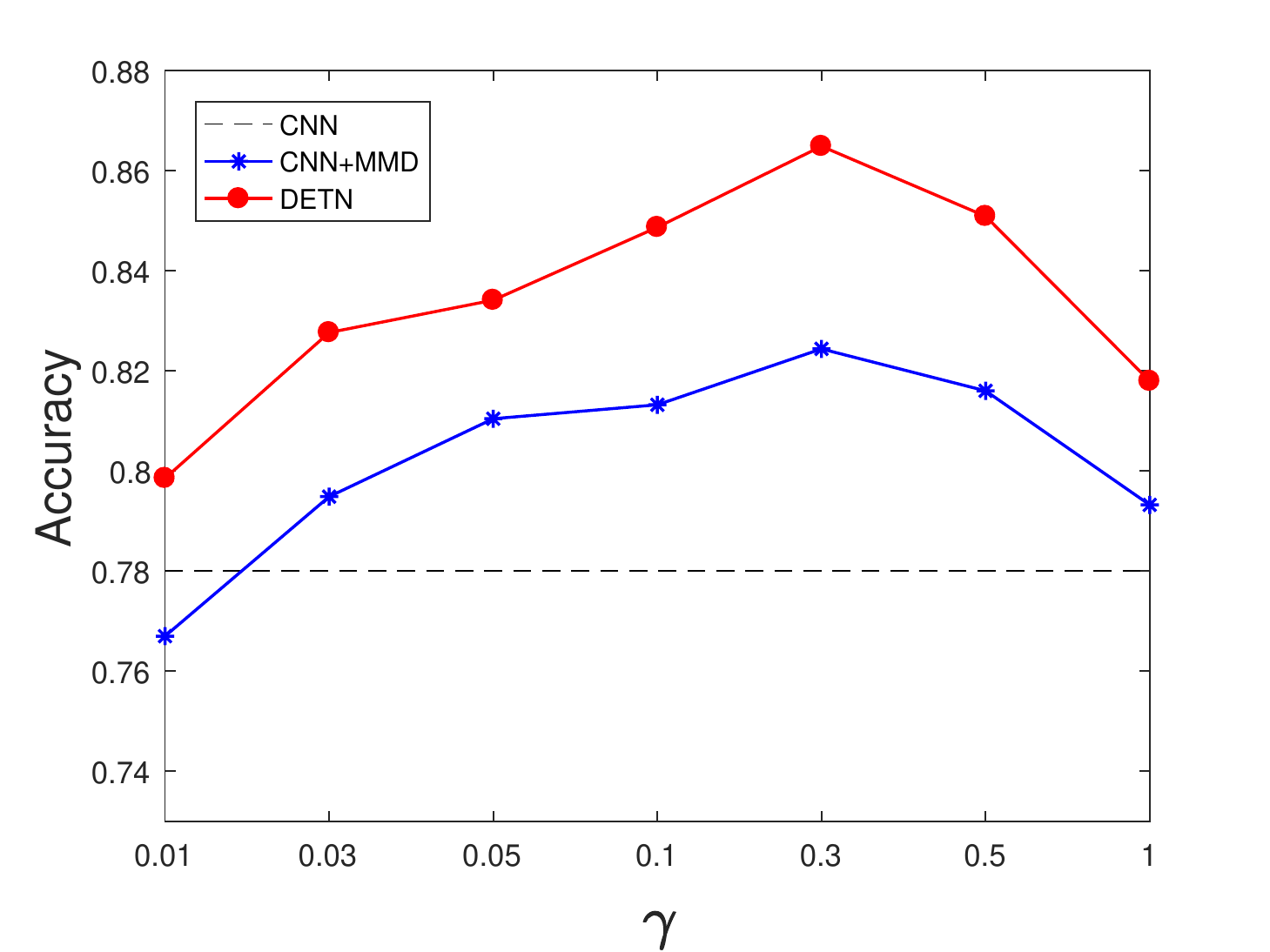}
			\includegraphics[width=2.75cm]{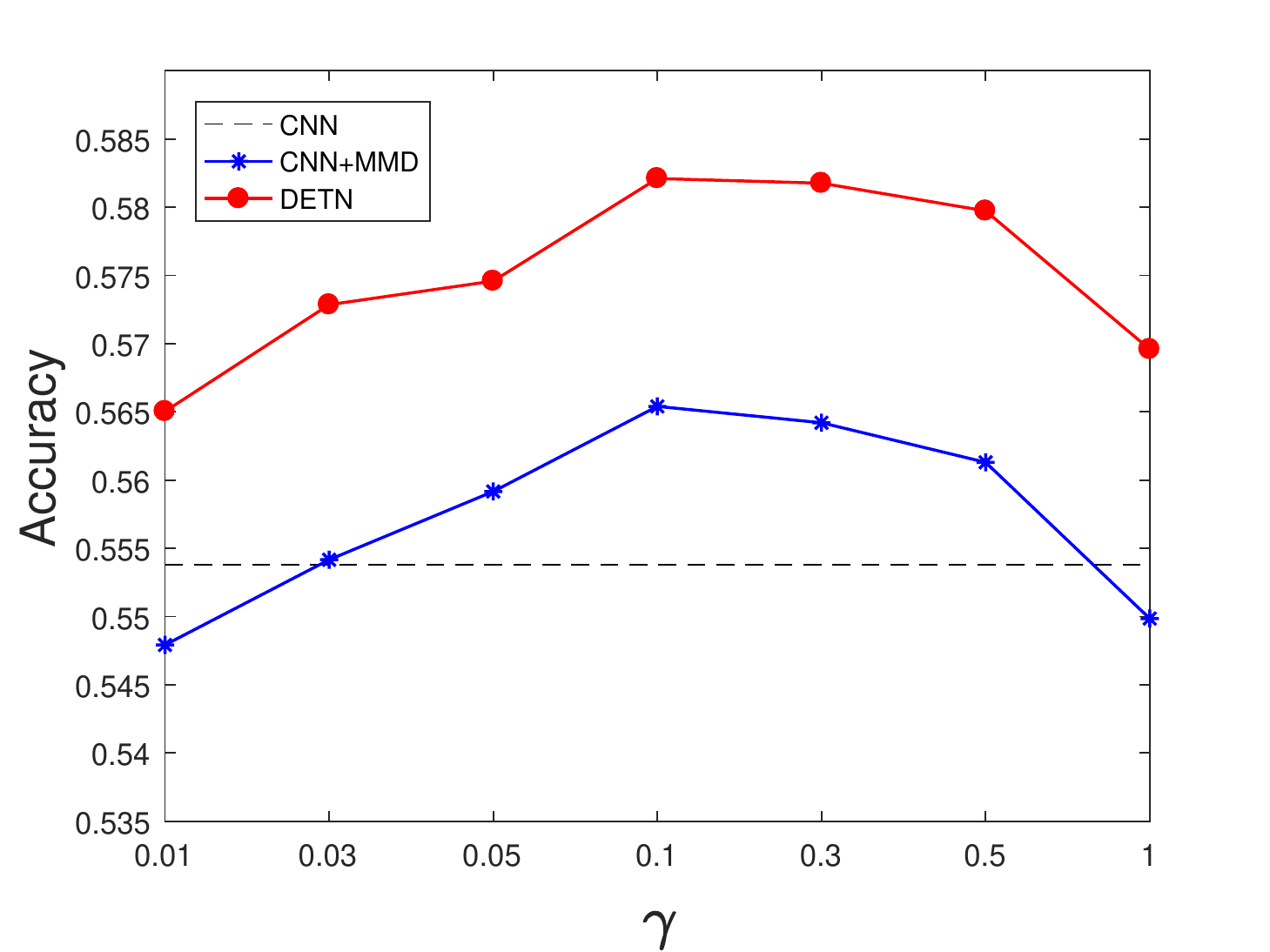}}
		\subfigure[\bm{$\lambda$}]{\label{par22}
			\includegraphics[width=2.8cm]{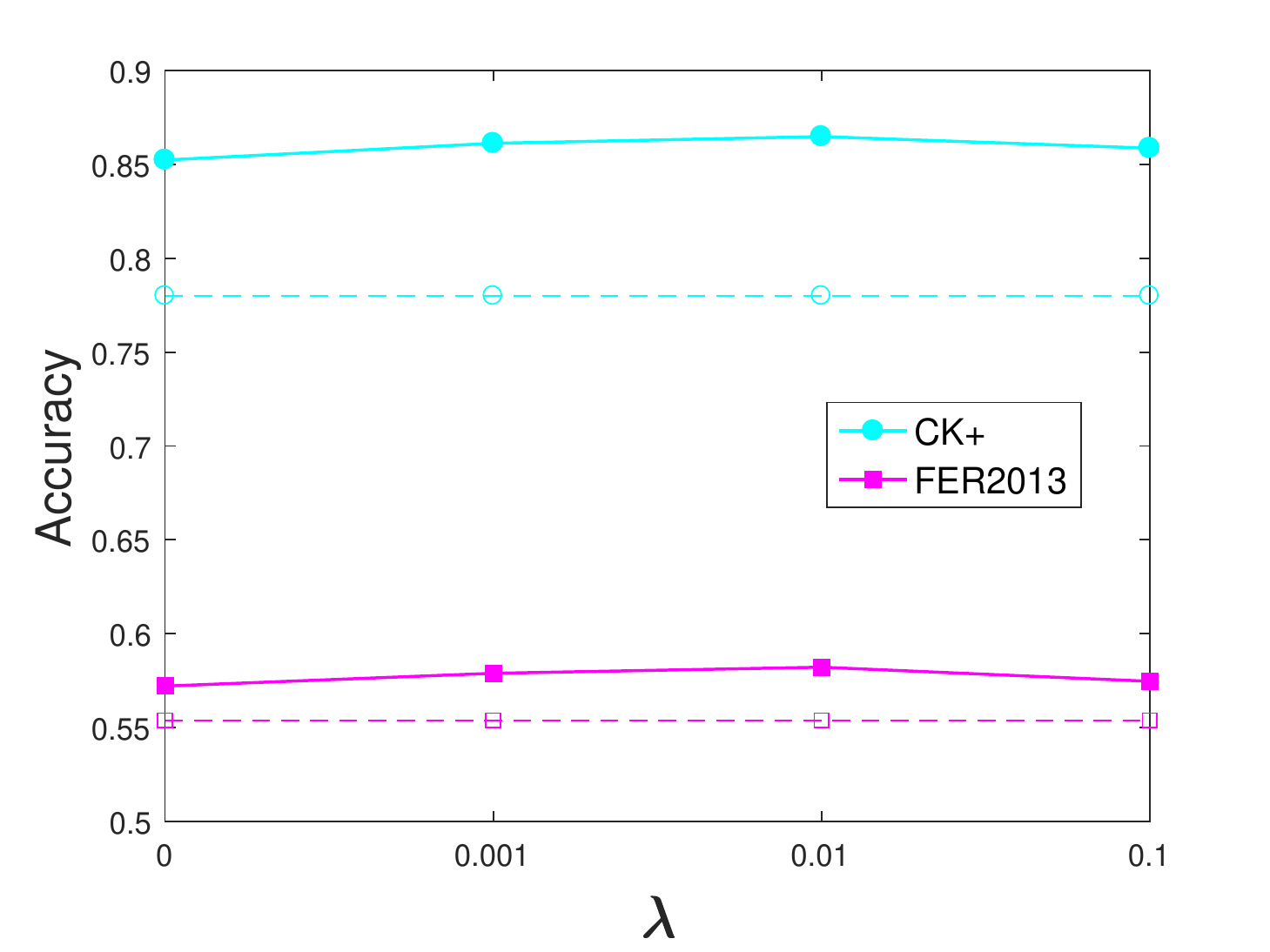}}
		\caption{Parameter sensitivity study on target datasets CK+ and FER2013 w.r.t $\gamma$ and $\lambda$. Results of the best baseline CNN are shown as the dashed lines.}
		\label{par}
	\end{figure}

	The tunable hype-parameters $\gamma$ and $\lambda$ in ECAN represent the domain transfer degree. To examine the sensitiveness of these two parameters, we demonstrate particular cross-domain results on the small-scale lab-controlled database CK+ and the large-scale in-the-wild database FER2013 in terms of different value of each hype-parameter, which can be seen in Fig. \ref{par}. The best results of the baseline CNN corresponding to the case $\gamma=0$ and $\lambda=0$ are reported in Fig. \ref{par} as the dashed lines.
	
	We first conduct experiment as $\gamma$ varying from 0.01 to 1, with $\lambda$ fixed. In Fig. \ref{par11}, it can be observed that for CK+ the accuracy reaches the maximum when $\lambda=0.3$ and then decreases as $\lambda$ continues to increase. Similarly, the performance on FER2013 peaks at $\lambda=0.1$ and then falls off. This bell-shaped curve exemplifies the promoting effect of the jointly supervision of the softmax loss and the MMD-based regularizers when a proper trade-off is chosen. Furthermore, when comparing the performances of our ECAN with the original MMD, our method behaves better than the original MMD as $\lambda$ varies. This suggests that the class imbalance indeed causes a bottleneck for facial expression recognition and the ECAN can effectively mitigate this problem by learning suitable re-sampling weights for the source domain.
	We then fix $\gamma$ and vary $\lambda$ from 0 to 0.1, and the results are shown in Fig. \ref{par22}. Note that $\lambda=0$ refers to the case of only using the re-weighted MMD term. And the further improvements as the increase of $\lambda$ indicates the importance and necessity of the conditional MMD term that matches the class conditional distributions and enhances the discriminative ability of the learned feature representations across domains in the meanwhile. 
	
	\subsubsection{Ablation study}
	To look more deeply into the proposed method, we compare several variants of ECAN: ``ECAN with re-weight'' that only considers the re-weighted MMD, ``ECAN with condition'' that only considers the conditional MMD and ``ECAN with re-weight and condition'' that considers both of two MMD regularizations.
	We then evaluate these variant algorithms on ``RAF-DB 2.0$\rightarrow$CK+''  cross-dataset task following the same experiment settings. The results are shown in Table \ref{tab:ablation}. From the ablation results, we can notice that all these variant methods of ECAN perform better than the baselines. And our ECAN that considers both marginal and conditional distributions and also the class distribution bias achieves the best result in these variants, which indicates that both two MMD regularization terms cooperate with each other to achieve desirable adaption behaviors.
	
	\begin{table}[t]
		\renewcommand\arraystretch{1.1}
		\setlength{\tabcolsep}{4.5pt}
		\centering
		\caption{Ablation study of ECAN on ``RAF-DB 2.0$\rightarrow$CK+''  cross-dataset facial expression recognition task.}
		\label{tab:ablation}
		\begin{tabular}{|c|c|c|c|}
			\hline
			Target&Methods&Source &Accuracy\\\hline
			\multirow{3}{*}{CK+}&ECAN (\textit{re-weight})&\multirow{3}{*}{\begin{tabular}[c]{@{}c@{}}RAF-DB\\ 2.0\end{tabular}}&85.24\%\\
			&ECAN (\textit{condition})&&83.56\%\\
			&ECAN (\textit{re-weight+condition})&&86.49\%\\\hline
		\end{tabular}
	\end{table}

	\subsubsection{Feature visualization}
\begin{figure}[t]
		\centering
		\subfigure[CNN+MMD]{\label{tsne1}
			\includegraphics[height=3.1cm]{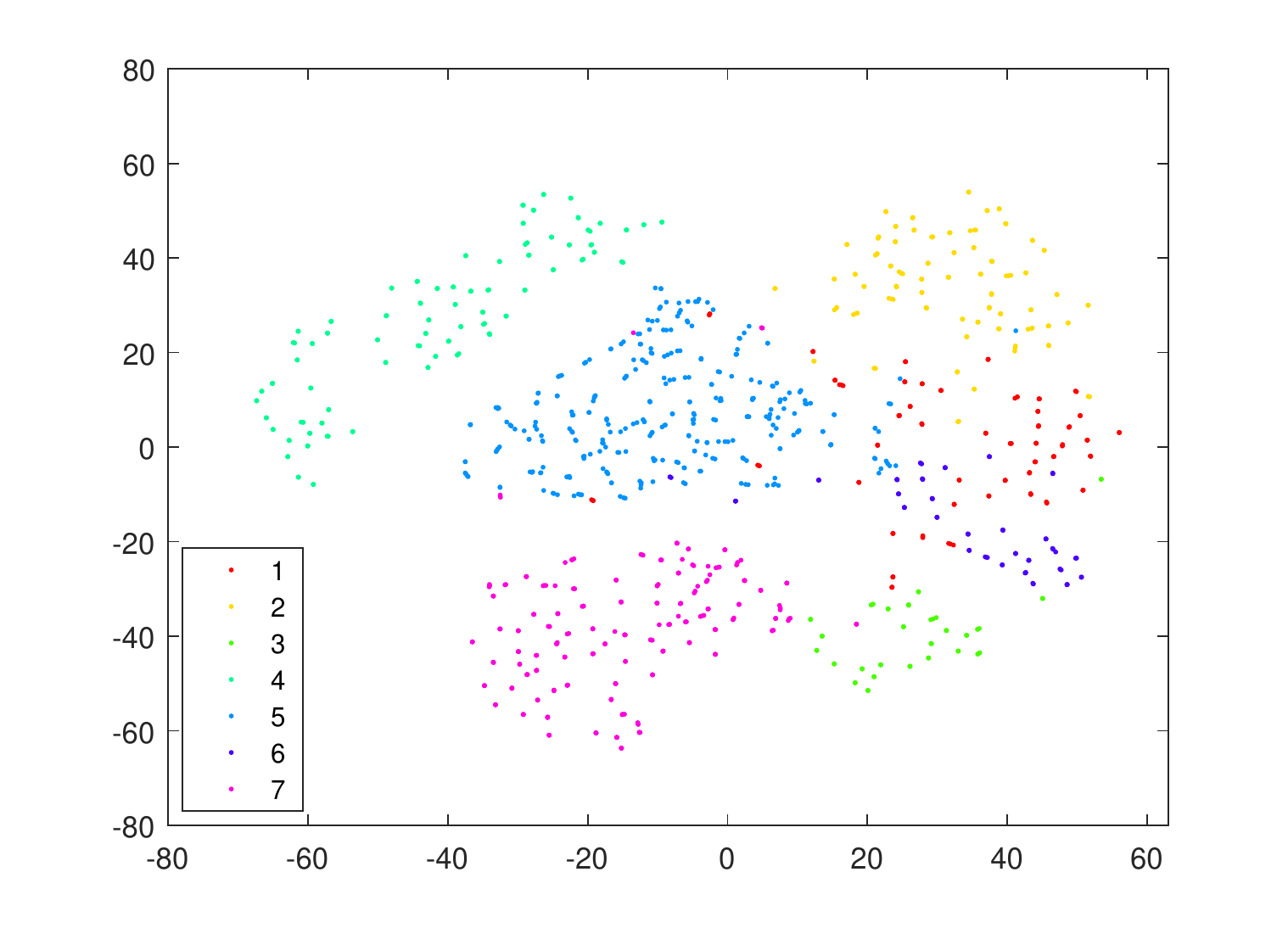}}
		\subfigure[ECAN]{\label{tsne2}
			\includegraphics[height=3.1cm]{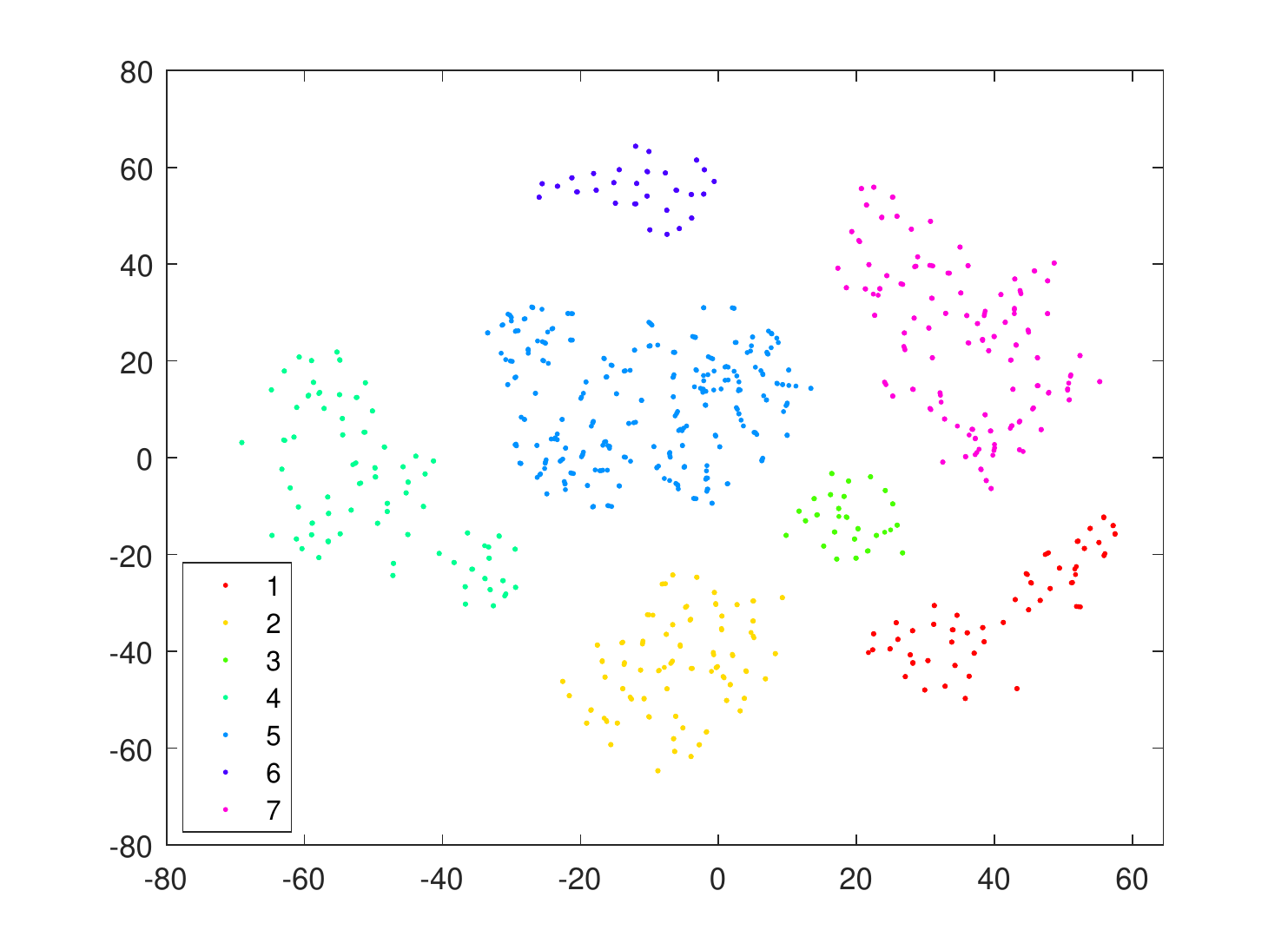}}
		\caption{Feature visualization results of different methods on CK+ database. 1, 2, 3, 4, 5, 6, 7 stand for Anger, Disgust, Fear, Happiness, Neutral, Sadness, Surprise respectively.}
		\label{tsne}
	\end{figure}

	To investigate how the ECAN works on cross-database facial expression recognition, we employed the t-SNE dimensionality reduction technique~\cite{van2014accelerating} to visualize  the learned features of CK+ on a 2-dimensional embedding. From Fig. \ref{tsne1} we can find that samples with anger and fear labels in CK+ tend to be misclassified into sadness and surprise respectively, which are two dominant categories in the source dataset RAF-DB 2.0. However, in Fig. \ref{tsne2}, our ECAN effectively alleviates this misguided decision caused by class distribution bias and decreases the classification error rate. What's more, with the help of class-conditional MMD term, samples with different labels can be pulled farther away, thus the separability and discriminative ability can be ensured.
	
	\section{Conclusion}
	\label{sec:conclusion}
	In this paper, a deep Emotion-Conditional Adaption Network (ECAN) has been proposed to conduct unsupervised cross-database facial expression recognition. 
	The strength of the ECAN lies in its ability to make the most of the beneficial knowledge from the target domain to simultaneously bridge the discrepancy of both marginal and conditional distribution between source and target domains, and also the discriminative power brought about by the learning process of deep network. Besides, class imbalance problem has been taken into account and a re-weighting parameter is introduced to balance the class bias between source and target domains. All these optimal goals associate with each other and then effectively boost the recognition rate of cross-database facial expression recognition. 
	Extensive experiments on widely-used facial expression datasets show that the proposed ECAN achieves excellent performance in a series of transfer tasks and outperforms previous cross-dataset facial expression recognition results, demonstrating the effectiveness of the proposed method.

	
	\ifCLASSOPTIONcaptionsoff
	\newpage
	\fi
	
	\bibliographystyle{IEEEtran}
	\bibliography{IEEEabrv,egbib}
	
	%
	

\end{document}